\documentclass[10pt,journal,compsoc]{IEEEtran}
\usepackage{hyperref}
\usepackage{url}
\usepackage{amsmath,amssymb}
\usepackage[dvips]{graphicx}
\usepackage{color,eucal,xspace}
\usepackage{algorithm}
\usepackage{algpseudocode}
\usepackage{pifont}

\graphicspath{{./}{./figs/}{../}{../figs/}}

\usepackage[caption=false,font=footnotesize,labelfont=sf,textfont=sf]{subfig}

\usepackage{fixltx2e}

\begin{document}
\title{Compositional Model Based Fisher Vector Coding for Image Classification}

\author{Lingqiao Liu, Peng Wang, Chunhua Shen,
  Lei Wang, Anton van den Hengel, Chao Wang, Heng Tao Shen
\IEEEcompsocitemizethanks{
  \IEEEcompsocthanksitem L. Liu, P. Wang, C. Shen and A. van den Hengel are with  School of Computer Science,
 The University of Adelaide,  SA, Australia.\protect\\
 E-mail: \{lingqiao.liu, peng.wang, chunhua.shen, anton.vandenhengel\}@\-a\-de\-lai\-de.\-edu.au

\IEEEcompsocthanksitem L. Wang and C. Wang are with School of Computing and Information Technology,
University of Wollongong, NSW, Australia.\protect\\
E-mail: \{leiw, chaow\}@uow.edu.au

\IEEEcompsocthanksitem
H. T. Shen is with School of Computer Science and Engineering, University of Electronic Science and Technology of China, Chengdu, China. \protect \\
E-mail: shenhengtao@hotmail.com

  }%
\thanks{}
}

\markboth{Appearing in IEEE Transactions on Pattern Analysis and Machine Intelligence, 01/2017}{Liu \MakeLowercase{\textit{et al.}}: }

\IEEEtitleabstractindextext{%
\begin{abstract}
Deriving from the gradient vector of a generative model of local features,
Fisher vector coding (FVC) has been identified as an effective coding method
for image classification. Most, if not all, FVC implementations employ the
Gaussian mixture model (GMM)
as the generative model for
local features. However, the representative power of a GMM can be limited because
it essentially assumes that local features can be characterized by a fixed
number of feature prototypes, and the number of prototypes is usually small in
FVC. To alleviate this limitation, in this work, we break the convention which
assumes that a local feature is drawn from one of a few Gaussian distributions.
Instead, we adopt a compositional mechanism which assumes that a local feature
is drawn from a Gaussian distribution whose mean vector
is composed as a
linear combination of multiple key components, and the combination weight is a
latent random variable. In doing so we greatly enhance the representative
power of the generative model underlying FVC.

To implement our idea, we design two
particular generative models
following this compositional approach.
In our first
model, the mean vector is sampled from the subspace spanned by a set of bases
and the combination weight is drawn from a Laplace distribution.  In our second
model, we further assume that a local feature is composed of a discriminative
part and a residual part.  As a result, a local feature is generated by the
linear combination of discriminative part bases and residual part bases. The
decomposition of the discriminative and residual parts is achieved via the
guidance of a pre-trained supervised coding method. By calculating the gradient
vector of the proposed models, we derive two new Fisher vector coding
strategies. The first is termed Sparse Coding-based Fisher Vector Coding
(SCFVC) and can be used as the substitute of traditional GMM based FVC. The
second is termed Hybrid Sparse Coding-based Fisher vector coding (HSCFVC)
since it combines the merits of both pre-trained supervised coding methods
and FVC. Using pre-trained Convolutional Neural Network (CNN) activations as
local features, we experimentally demonstrate that the proposed methods are
superior to traditional GMM based FVC and  achieve  state-of-the-art
performance in various image classification tasks.
\end{abstract}
}
\maketitle

\IEEEdisplaynotcompsoctitleabstractindextext
\clearpage
\tableofcontents
\clearpage

\section{Introduction}
In the bag-of-features model, Fisher vector coding \cite{FV_First,ImprovedFV} (FVC) is a coding method derived from the Fisher kernel \cite{FisherKernelOriginal} which was originally proposed to compare two samples induced by a generative model. The basic idea of FVC is to first construct a generative model of local features and
use the gradient of the log-likelihood of a particular feature with respect to the model parameters as the feature's coding vector.
When applied
as an image representation
the FVC vectors of local features are
calculated
by a pooling operation and
normalization \cite{ImprovedFV}
to generate the final image representation.
FVC has been established as one of the most powerful local feature encoding and image representation generation methods.
In most of the visual classification systems with FVC, Gaussian mixture model (GMM) is adopted as the generative model for modeling the local features. The GMM essentially assumes that each local feature is generated from one of the Gaussian distributions in the GMM, and intuitively the mean of each Gaussian distribution serves as a prototype for the local features. Since the dimensionality of the image representation resulting from GMM based FVC is the product of the local feature dimensionality and the number of Gaussians, to make the image representation dimensionality tractable, the number of Gaussians is usually chosen to be few hundred.

With the recent development in feature learning \cite{CNN_Baseline},  higher dimensional local features such as the activations of a pre-trained deep neural network \cite{CNN_Regional,TextureFV,Our_NIPS,SemanticFisherVector} have become increasingly popular. However, modeling these local features with the GMM for FVC
is
challenging. This is due to two factors:
1) The dimensionality of these local features can be much higher than that of the traditional local features, e.g., SIFT. As a result, the feature space spanned by these local features can be very large and using limited number of Gausssian distributions can be insufficient to accurately model the true feature distribution.
2) The number of Gaussian distributions cannot be large due to the resulting increase in the local feature dimensionality
and the corresponding increase in the size of the image-level representation.

To tackle the challenge of using high-dimensional local features in FVC, we propose two alternative solutions in building the generative model. Both solutions rely on the idea of compositional modeling which assumes that \textit{a local feature can be better modeled as the composition of multiple components  than by using a prototype}. For many recently proposed local features, such as CNN activations on local image regions, the image area that a local feature covers is relatively large. In this case, compositional modeling is a more natural choice than single prototype modeling because the visual pattern within the local region is clearly a combination of multiple object/scene parts. Mathematically, we formulate the aforementioned idea as a two-stage generative process: in the first stage, the combination coefficients of multiple bases are drawn from a distribution and a linear combination of bases is generated; in the second stage, a local feature is drawn from a Gaussian distribution whose mean vector is the combined vector generated from the first stage. The compositional components in the proposed methods are treated as model parameters which are learned subsequently.

Two encoding approaches are proposed in this paper.
The main difference between the two proposed approaches lies
at the ways
of decomposing a local feature.
The first approach adopts a single basis matrix and assumes that
each combination coefficient is drawn from a Laplace
distribution. The second approach
takes the further step, by assuming
that a local feature may be decomposed into a discriminative part and a residual part.
The discriminative part represents those patterns which are found to be discriminative and the residual part depicts the patterns which are not well captured by the identified discriminative part. To achieve such a decomposition, we rely on a pre-trained supervised coding method and use its coding vector as
our guide.
The motivation for using decomposition-based modeling is twofold:
\begin{itemize}
  \item
This decomposition  enables part of the generative model to focus more on the discriminative part and
thus  better captures class-specific information.
\item
  On the other hand, the discriminative part identified by the pre-trained supervised coding method may not be able to
  capture all the useful patterns in the local features due to the imperfection of supervised encoder training\footnote{This
may be due to poor
local minima caused by training of a non-convex objective function, or the overfitting phenomenon due to the difficulty of regularizing
a deeply trained supervised encoder.}.
In this case, the part of the generative model which models the residual provides a second chance to distill the missing information and thus compensates for the discriminative part modeling.
\end{itemize}
Due to the complementary nature
of the discriminative and residual parts, as well as the high dimensionality of Fisher vectors, it is expected that the Fisher vector derived from our second model  preserves more useful information than our first FVC and the supervised coding method that guides the decomposition.

Moreover, we show that, under some certain approximation,
the inference and learning problems of both methods can be converted into
variants of  sparse-coding problems which
can be readily solved using
an off-the-shelf sparse coding solver. For this reason, we name the FVC derived from the first and the second models
as Sparse Coding-based Fisher Vector Coding (SCFVC) and Hybrid Sparse Coding-based Fisher Vector Coding (HSCFVC).

To accelerate the calculation, we also develop efficient approximation solutions based on the matching pursuit algorithm
\cite{MatchingPursuit}. By conducting intensive experimental evaluation on object classification, scene classification, and fine-grained image classification problems, we demonstrate that the proposed methods are superior to the traditional GMM-based FVC. HSCFVC further demonstrates state-of-the-art classification performance on  evaluated benchmark datasets.

A preliminary version of the first proposed method was published in \cite{Our_NIPS}. In this paper we
have extended the approach significantly,
and in particular we develop HSCFVC which generalizes the framework of SCFVC and leads to further improved classification performance.

\section{Related work}
In this section, we briefly review some relevant work.
\subsection{Fisher vector coding}
The concept of Fisher vectors was originally proposed in \cite{FisherKernelOriginal} as a framework to build a discriminative classifier from a generative model. It was later applied to image classification \cite{FV_First} by modeling the image as a bag of local features sampled from an i.i.d.\ distribution. Later, several variants were proposed to improve the  original  FVC.
One of the first identified facts is that normalization
of Fisher vectors is essential for achieving good performance \cite{ImprovedFV}.
At the same time, several similar variants were developed independently from different perspectives
\cite{VLAD,SuperVector_ECCV,SuperVector}. The improved Fisher vector and  variants showed state-of-the-art performance
in image classification and quickly became one of the most popular visual representation methods in computer vision.
Numerous approaches have been developed to further enhance  performance. For example, The work in \cite{All_about_VLAD}
closely analyzed particular implementation details of VLAD, a famous  variant of FVC.
The work in \cite{FisherLayout} attempted to incorporate  spatial information from local features into the Fisher vector framework.
In \cite{NonIID_FV,NonIID_FV_PAMI}, the authors revisited the basic i.i.d.\ assumption of FVC and pointed out its limitation.
They proposed a non-iid model and derived an approximated Fisher vector encoding method for image classification.
Furthermore, FVC has been widely applied to various vision applications and has demonstrated state-of-the-art performance in
those areas. For example, in combination with local trajectory features, FVC-based systems have achieved  state-of-the-art
performance in video-based action recognition \cite{ImprovedTrajectory,StackedFV}.

\subsection{FVC with CNN local features}
Conventionally, most FVC implementations are applied to low-dimensional hand-crafted local features, such as SIFT \cite{SIFT}. With the recent development of deep learning, it has been observed that simply extracting neural activations from a pre-trained CNN model
achieves
significantly better performance \cite{CNN_Baseline}. However, it was soon
discovered that directly using activations from a pre-trained CNN as global features may not be
the optimal choice \cite{CNN_Regional,TextureFV,Our_NIPS,SemanticFisherVector}, at least for
small/medium sized classification problems for which fine-tuning a CNN does not always improve performance significantly.
Instead, it has been shown that it is beneficial to view CNN activations as local features.
In this case, the traditional local feature coding approaches, such as FVC, can be readily applied. The work in \cite{CNN_Regional}
points out that the fully-connected activation of a pre-trained CNN is not translation invariant. Thus, the authors proposed
to extract CNN activations from multiple regions of an image and use VLAD to encode these local features.
In \cite{TextureFV} and \cite{CrossLayerPooling}, the value of convolutional layer activations are analyzed.
They suggested that convolutional feature activations can be seen as a set of local features extracted at a dense grid.
In particular, the work in \cite{TextureFV} builds a texture classification system by applying FVC to the convolutional layer features
of a CNN.

\subsection{Supervised coding and FVC}\label{sect:supervised encoding related work}
The proposed HSCFVC combines the idea of supervised coding and FVC. Here we briefly review the work of supervised coding and the attempts to combine it with FVC. Using  discriminative information to create an image representation is a popular idea in image classification. For example, label-supervised information has been used
to learn discriminative codebooks for encoding local features \cite{Info_SupCodebook,max_margin_dictionary_learning,SupKSVD,Yang10suptrans-invariant,MaxmarginMIDL}, either by using a separated codebook learning step \cite{Info_SupCodebook,SupKSVD} or in an end-to-end fashion \cite{max_margin_dictionary_learning,Yang10suptrans-invariant}. Supervision information has also been applied to discover a set of
middle-level discriminative patches \cite{Mid-LevelNIPS,BlocksThatShout,Yao_Mining}
to train some patch detectors which are essentially local feature encoders.
A CNN can also be seen as a special case of supervised coding methods if we view the responses of the filter bank in a convolutional layer
as the coding vector of convolutional activations of the previous layer.
From this perspective, a deep CNN can be seen as a hierarchical extension of  supervised coding methods.

Generally speaking, the aforementioned supervised coding and FVC represent two major methodologies for creating discriminative
image representations. For supervised coding, the supervision information is passed through the early stage of a classification system, namely,
by learning a dictionary or coding function.
 For FVC, the information  of local features will be largely preserved
 in the corresponding high-dimensional signature. Then a simple classifier can be used to extract the discriminative patterns
 for the final classification. There have been several works trying to combine the idea of FVC and supervised coding.
 The work in \cite{deep_fisher_kernel} learns the model parameters of FVC in an end-to-end supervised training framework.
 In \cite{End_to_end_deep_fisher}, multiple layers of Fisher vector coding modules are stacked into a deep architecture to form a deeper network.
 In contrast to these works, our HSCFVC is based on the basic conceptual framework of FVC:
 First we build a generative model and then derive its gradient vector. %

\section{Background}
Before we present our methods, we give an introduction to the standard Fisher vector coding method.

\subsection{Fisher vector coding}
Given two samples generated from a generative model, their similarity can be evaluated by the Fisher kernel \cite{FisherKernelOriginal}. The samples can take any form, including a vector or a vector set, as long as its generation process can be modeled. For the Fisher vector-based image classification approach, the sample is a set of local features extracted from an image which we denote as $\mathbf{X} = \{\mathbf{x}_1,\mathbf{x}_2,\cdots,\mathbf{x}_T\}$. Assuming that $\mathbf{x}_i$ is drawn i.i.d.\ from the distribution $P(\mathbf{x}|\lambda)$, in the Fisher kernel a sample $\mathbf{X}$ can be described by the gradient vector of the likelihood function w.r.t.\ the model parameter $\lambda$
\begin{align}\label{eq:FisherKernel}
	\mathbf{G}_{\lambda}^{\mathbf{X}} = \nabla_{\lambda} \log P(\mathbf{X}|\lambda) = \sum_i \nabla_{\lambda}\log P(\mathbf{x}_i|\lambda).
\end{align}
The Fisher kernel is then defined as $K(\mathbf{X},\mathbf{Y}) = {\mathbf{G}_{\lambda}^{\mathbf{X}}}^T \mathbf{F}^{-1} \mathbf{G}_{\lambda}^{\mathbf{Y}}$, where $\mathbf{F}$ is called information matrix and is defined as $\mathbf{F}= E[\mathbf{G}_{\lambda}^{\mathbf{X}}{\mathbf{G}_{\lambda}^{\mathbf{X}}}^T]$.  \textcolor{black}{In this paper, we follow \cite{FisherKernelOriginal} to omit it for computational simplicity. However, we can also approximate it by whitening the dimensions of the gradient vector $\mathbf{G}_{\lambda}$ as suggested in \cite{ImprovedFV}.}
As a result, two samples can be directly compared by the linear kernel of their corresponding gradient vectors which are often called Fisher vectors. From a bag-of-features model perspective, the evaluation of the Fisher kernel for two images can be seen as first calculating the gradient or Fisher vector of each local feature and then performing sum-pooling. In this sense, the Fisher vector of each local feature, $\nabla_{\lambda}\log P(\mathbf{x}_i|\lambda)$, can be seen as a coding vector and we call it Fisher vector coding in this paper.

\subsection{Gaussian mixture model-based FVC}%
To implement the Fisher vector coding framework introduced above, one needs to specify the distribution $P(\mathbf{x}|\lambda)$. In the literature, most works use a GMM to model the generation process of $\mathbf{x}$, which can be described as follows:
\begin{itemize}
	\item Draw a Gaussian model $\mathcal{N}(\mu_k,\Sigma_k)$ from the prior distribution $P(k),~k=1,2,\cdots,m$ .
	\item Draw a local feature $\mathbf{x}$ from $\mathcal{N}(\mu_k,\Sigma_k)$.
\end{itemize}
Generally speaking, the distribution of $\mathbf{x}$ resembles a Gaussian distribution only within a local region of the feature space. Thus for a GMM, each Gaussian distribution in the mixture only models a small partition of the feature space and intuitively each Gaussian distribution can be seen as a feature prototype. As a result, a number of Gaussian distributions will be needed to accurately depict the whole feature space.  For commonly used low dimensional local features, such as SIFT \cite{SIFT}, it has been shown that it is sufficient to set
the number of Gaussian distributions to be of the order of a few hundred. However, for higher dimensional local features this number may be insufficient. This is because the volume of feature space usually increases quickly with the feature dimensionality. Consequently, the same number of Gaussian distributions will leave a coarser partition resolution and may lead to imprecise modeling.

To increase the partition resolution for higher dimensional feature spaces, one straightforward solution is to increase the number of Gaussian distributions. However, it turns out that the partition resolution increases slowly (compared to our method which will be introduced in the next section) with the number of Gaussian distributions. In other words, much larger numbers of Gaussian distributions will be needed and this will result in a Fisher vector whose dimensionality is too high to handle in practice.

\section{Our approaches}
\subsection{Compositional generative model}
Our solution to this issue is to adopt a compositional model which does not model local features via a fixed number of prototypes. Instead, it assumes that the prototype can be adaptively generated by the composition of multiple pre-learned components. In other words, we can essentially leverage an infinite number of prototypes to model the whole feature space. Thus the representative power of the generative model can be substantially improved. Intuitively, our model is motivated by the fact that many visual patterns within a local image region, especially those in a relatively large local region, can be seen as the combination of multiple object or scene parts. The complexity of those visual patterns can be attributed to the large number of possible combinations of some elementary patterns. So it is more efficient to use those elementary patterns to model the visual patterns rather than
to attempt to directly model all
possible pattern combinations.

Based on this insight, in this work we propose a two-stage framework to model the generative process of a local feature, which can be expressed as follows:
\begin{itemize}
	\item Draw a latent combination coefficient $\mathbf{u}$ from a pre-specified distribution $P(\mathbf{u})$.
	\item Generate a prototype $\mathbf{\mu}$ by linearly combining the elementary patterns $\mathbf{B}$ with the latent combination coefficient, that is, $\mathbf{\mu} = \mathbf{B}\mathbf{u}$. Then draw a local feature from the Gaussian distribution $\mathcal{N}(\mathbf{\mu},\Sigma)$.
\end{itemize}
In this model, $\mathbf{B} \in \mathbb{R}^{d\times m}$ denotes $m$ elementary patterns and is treated as the model parameters. Also note that in this framework, we do not treat the mean vector $\mathbf{\mu}$ as the model parameter but as a mapping from the latent combination coefficient. Thus we can essentially generate the infinite number of Gaussian distributions by varying $\mathbf{u}$. By doing so, we can significantly increase the representative power of the generative model while keeping the number of its parameters, which determines the dimensionality of the resulted Fisher vector, being tractable.

One question remains that is how to model $P(\mathbf{u})$, the distribution of the latent combination coefficient. In this work, we propose two different ways to model this distribution.

\subsubsection{Approach I (SCFVC)}
The first approach models $P(\mathbf{u})$ as a Laplace distribution. In other words, it assumes that the combination weight is sparse. This choice follows the common belief that visual signals can be modeled by the sparse combination of over-complete bases. Once the combination coefficient is sampled, we generate the prototype $\mu$ via $\mathbf{B}\mathbf{u}$. More specifically, the generative process is written as follows:
\begin{itemize}
	\item Draw a coding vector $\mathbf{u}$ from a zero mean Laplace distribution $P(\mathbf{u}) = \frac{1}{2\lambda}\exp(-\frac{|\mathbf{u}|}{\lambda})$.
  \item Draw a local feature $\mathbf{x}$ from the Gaussian distribution $\mathcal{N}(\mathbf{B}\mathbf{u},\Sigma)$.
\end{itemize}
Note that the above process resembles a sparse coding model. To show this relationship, let us first write the marginal distribution of $\mathbf{x}$ according to the above generative process:
\begin{align}
	P(\mathbf{x}) = \int_{\mathbf{u}} P(\mathbf{x},\mathbf{u}|\mathbf{B}) d \mathbf{u} = \int_{\mathbf{u}} P(\mathbf{x}|\mathbf{u},\mathbf{B}) P(\mathbf{u}) d \mathbf{u}.
\end{align}
The above formulation involves an integral operator which makes the likelihood evaluation difficult. To simplify the calculation, we use the point-wise maximum within the integral term to approximate the likelihood\footnote{\textcolor{black}{Strictly speaking,
due to this approximation the resulting descriptors do not exactly correspond to Fisher kernels. Instead they are Fisher vector-like encoding methods.}}, that is,
\begin{align}
	 P(\mathbf{x}) &\approx P(\mathbf{x}|\mathbf{u^*},\mathbf{B}) P(\mathbf{u^*}). \nonumber \\
   \mathbf{u^*} &= \mathop{\mathrm{argmax}}_{\mathbf{u}} P(\mathbf{x}|\mathbf{u},\mathbf{B}) P(\mathbf{u}).
\end{align}
By assuming that $\Sigma = diag(\sigma_1^2,\cdots,\sigma_m^2)$ and set $\sigma_1^2 = \cdots = \sigma_m^2 = \sigma^2$ as a constant, the negative logarithm of $P(\mathbf{x})$ is written as
\begin{align}\label{eq:SCFVC_Log}
-\log(P(\mathbf{x}|\mathbf{B})) = \min_{\mathbf{u}} \frac{1}{\sigma^2} \|\mathbf{x}- \mathbf{B}\mathbf{u}\|^2_2 + \lambda \|\mathbf{u}\|_1, \nonumber
\end{align}
which is exactly the objective value of a sparse coding problem. This relationship suggests that we can learn the model parameter $\mathbf{B}$ and infer the latent variable $\mathbf{u}$ by using off-the-shelf sparse coding solvers.

\begin{figure*}[!ht]
	\centering
    \begin{tabular}{l}
            \subfloat[]{ \includegraphics[height=45mm,width=55mm ]{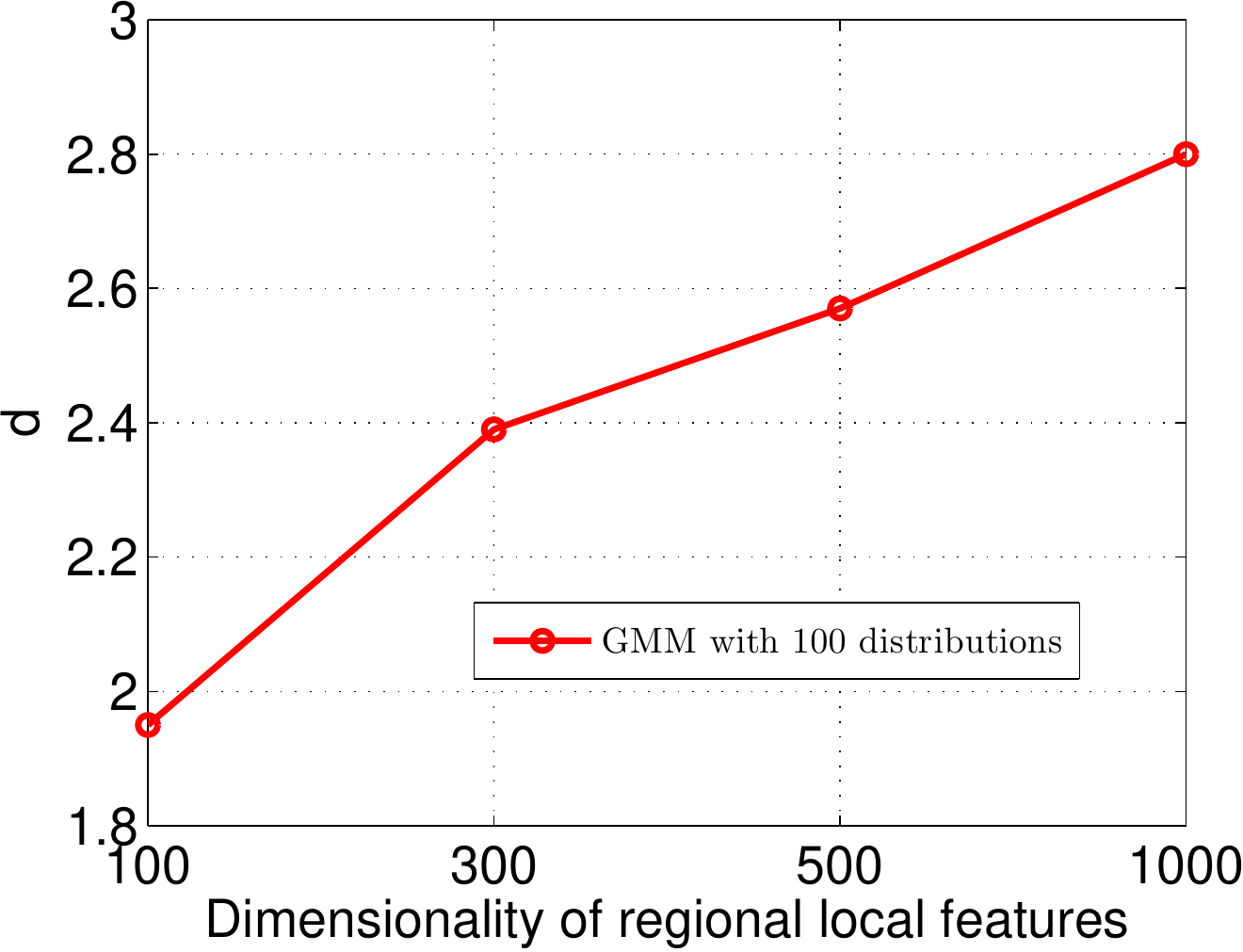}}
            \subfloat[]{ \includegraphics[height=45mm,width=55mm ]{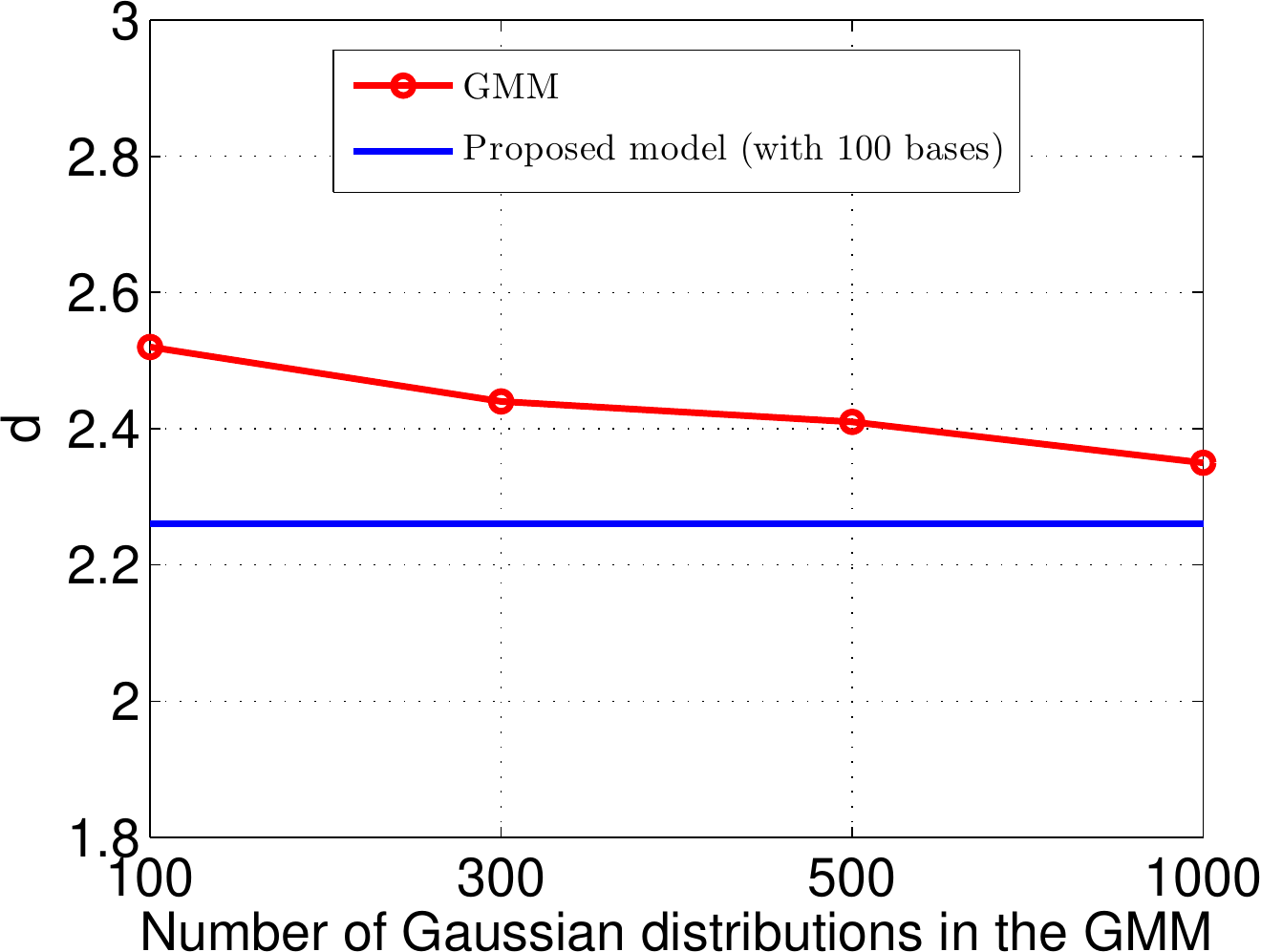}} \\
    \end{tabular}
    \caption{\textcolor{black}{Comparison of two strategies to increase the modeling accuracy. (a) For GMM, $d$, the average distance (over 500 sampled local features) between a local feature and its closest mean vector, increases with the local feature dimensionality with the number of GMM is fixed at 100. (b) $d$ is reduced by two ideas (1) simply increasing the number of Gaussian mixtures. (2) using the proposed generation process. As we see, the latter achieves much lower $d$ even with a small number of bases.} }
    \label{fig:compare_partition_resolution}
\end{figure*}

An  obvious question with respect  the method described above is whether it improves modeling accuracy significantly
over simply increasing the number of Gaussian distributions of the traditional GMM.
To answer this question, we design an experiment to compare these two schemes. In our experiment,
we use the average distance (denoted by $\mathrm{d}$) between a feature and its closest mean vector in the GMM or
the above model as the measurement for modeling accuracy. The larger $\mathrm{d}$, the lower the accuracy is.
The comparison is shown in Figure \ref{fig:compare_partition_resolution}. In Figure \ref{fig:compare_partition_resolution}(a),
we increase the dimensionality of local features\footnote{This is achieved by performing PCA on a 4096-dimensional CNN regional descriptor. For more details about the feature used,  refer to Section \ref{sect:deepCNN_Feature}.}
and for each dimensionality we calculate $\mathrm{d}$ in a GMM model with 100 Gaussian distributions. As can be seen, $\mathrm{d}$ increases quickly with the feature dimensionality.
In Figure \ref{fig:compare_partition_resolution}(b), we see that it is possible to reduce $\mathrm{d}$ by introducing more Gaussian  distributions into the GMM model. However, as may be seen, $\mathrm{d}$ drops slowly with the increase of the number of mixtures. In contrast, with the proposed method, we can achieve much lower $\mathrm{d}$ using only 100 bases. This result demonstrates the motivation of our method.

\begin{figure}[ht]
\begin{center}
\includegraphics[scale=0.3]{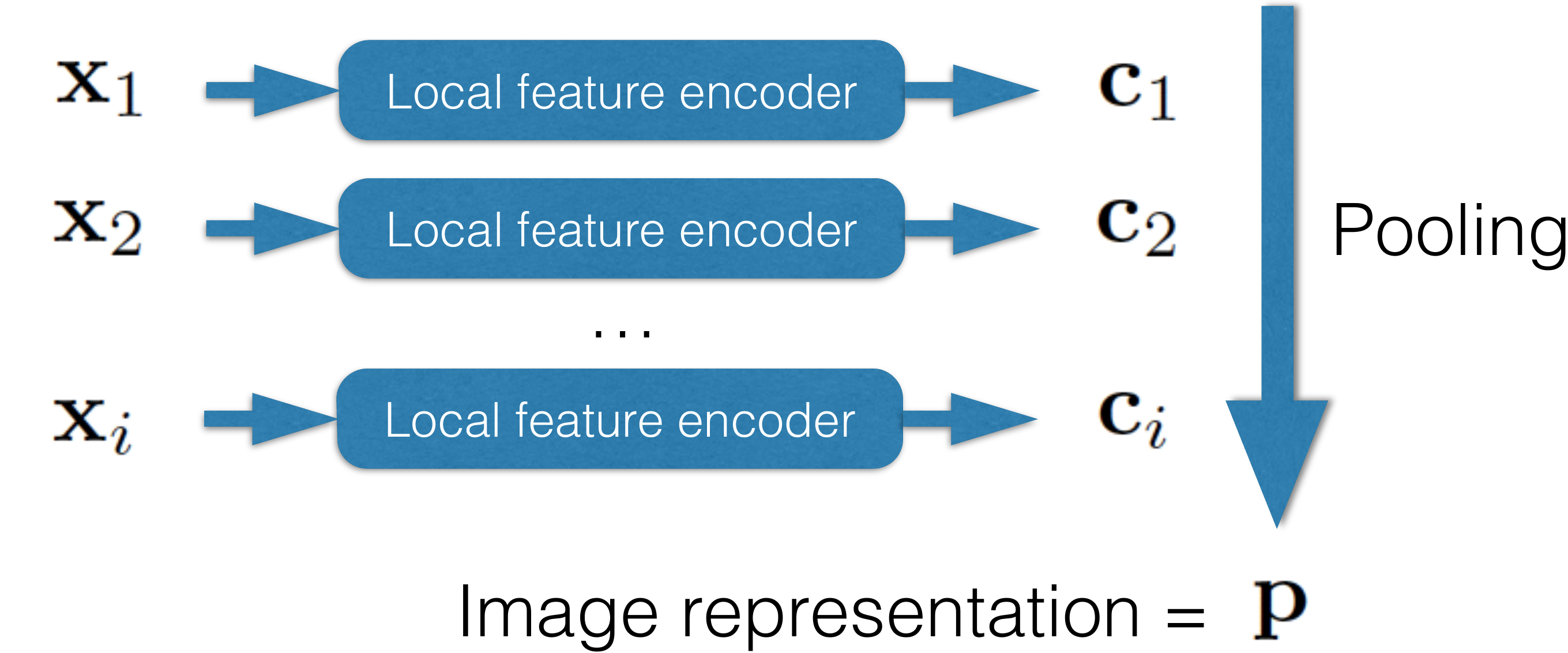}
\end{center}
\caption{\textcolor{black}{The demonstration of the supervised coding method. In a supervised coding method, the supervision information is used to learn the encoder function. A supervised coding method is used to guide the decomposition of the discriminative part and the residual part of a local feature.}}
\label{fig_encoder_demo}
\end{figure}

\subsubsection{Approach II (HSCFVC)}
The second approach that we propose for modeling $P(\mathbf{u})$ is based on a further decomposition of the local feature. In this approach, a local feature is assumed to be composed of a discriminative part and a residual part:
\begin{align}
	\mathbf{x} = \mathbf{x}_d + \mathbf{x}_r,
\end{align}
where $\mathbf{x}_d$ and $\mathbf{x}_r$ denote the discriminative part and the residual part respectively. The discriminative part indicates the visual pattern that is identified as informative for discrimination by an oracle method. The residual part in this decomposition can either correspond to the patterns shared by many classes, the irrelevant visual patterns or the remaining useful information which has not been successfully identified by the oracle method.
The motivation for modeling these two components separately is that they offer different values for the final application,
and modeling them jointly thus may undermine the discriminative power of the resulting Fisher vector.
The problem of how to achieve this decomposition remains, however.
Clearly, there are infinitely possibilities to decompose $\mathbf{x}$ into $\mathbf{x}_d$ and $\mathbf{x}_r$. To solve this problem, we resort to the guidance of a pre-trained supervised coding method (we will discuss the specific choice in Section \ref{sect:supervised_encoding}). The idea of the supervised coding method is demonstrated in Fig. \ref{fig_encoder_demo}, the supervised coding method maps each local feature $\mathbf{x}$ to a coding vector $\mathbf{c}$ and pools coding vectors from all local features to obtain the image-level representation. It encompasses a wide range of feature coding methods, such as those discussed in Section \ref{sect:supervised encoding related work}. In this paper we further assume that $\mathbf{c}$ is sparse. This is a reasonable assumption since many supervised encoding methods explicitly enforce the sparsity property \cite{SupKSVD,Yang10suptrans-invariant} and the coding vectors from many other methods can be sparsified by thresholding \cite{Mid-LevelNIPS} or simply setting top-$k$ largest coding values to be nonzero \cite{Yao_Mining}. For those kinds of supervised coding methods, the presence of a nonzero coding value essentially indicates the occurrence of a discriminative elementary pattern identified by the supervised coding method. In other words, each active (non-zero) coding dimension corresponds to one discriminative elementary pattern and the discriminative part of the local feature is the combination of these patterns. Let $\mathbf{B}_d$ denote the collection of discriminative elementary patterns (bases) and $\mathbf{u}_d$ be their corresponding combination weight. The above insight motivates us to encourage $\mathbf{u}_d$ to share the similar nonzero dimensions with $c$ , that is, to require $\|\mathbf{u}_d-\mathbf{c}\|_0$ to be small. However, the $l_0$ norm makes the Fisher vector derivation difficult. Thus we relax $l_0$ norm to $l_2$ norm in our approach.

To incorporate the above ideas into our two-stage feature generative process framework, we assume that $\mathbf{x}_d$ and $\mathbf{x}_r$ are drawn from Gaussian distributions whose mean vectors are the linear combination of two bases $\mathbf{B}_d$ and $\mathbf{B}_r$ respectively. For the combination weight of the residual part $\mathbf{u}_r$, we still assume that it is drawn from a Laplace distribution. The combination weight of the discriminative part $\mathbf{u}_d$, however is assumed drawn from a compound distribution which should encourage both sparsity and compatibility with the supervised coding $\mathbf{c}$. More specifically, we propose the following generative process of $\mathbf{x}$:
\begin{itemize}
	\item Draw a coding vector $\mathbf{u}_d$ from the conditional distribution $P(\mathbf{u}_d|\mathbf{c})$.
	\item Draw a coding vector $\mathbf{u}_r$ from a zero mean Laplace distribution $P(\mathbf{u}_r) = \frac{1}{2\lambda}\exp(-\frac{\|\mathbf{u}_r\|_1}{\lambda_1})$.
	\item Draw a local feature $\mathbf{x}$ from the Gaussian distribution $\mathcal{N}(\mathbf{B}_d\mathbf{u}_d + \mathbf{B}_r\mathbf{u}_r,\Sigma)$, where $\mathbf{B}_d$ and $\mathbf{B}_r$ are model parameters. Here we define $\Sigma = diag(\sigma_1^2,\cdots,\sigma_m^2)$ and set $\sigma_1^2 = \cdots = \sigma_m^2 = \sigma^2$ as a constant.
\end{itemize}
In the above process, $P(\mathbf{u}_d|\mathbf{c})$ is defined as $\frac{1}{Z} \exp\left(-\frac{\|\mathbf{u}_d\|_1}{\lambda_2}  -\frac{\|\mathbf{u}_d-\mathbf{c}\|_2}{\lambda_3}\right)$ to meet its two requirements as discussed above, where
\[
  Z = \int\limits_{\mathbf{u}_d} \exp\left(-\frac{\|\mathbf{u}_d\|_1}{\lambda_2}  -\frac{\|\mathbf{u}_d-\mathbf{c}\|_2}{\lambda_3}\right)\mathrm{d}\mathbf{u}_d
\]
is a constant. Also note that we do not separately generate the discriminative and common part of $\mathbf{x}$ in practise, i.e.,
$\mathbf{x}_d\sim \mathcal{N}(\mathbf{B}_d\mathbf{u}_d,\bar{\Sigma})$, $\mathbf{x}_r\sim \mathcal{N}(\mathbf{B}_r\mathbf{u}_r,\bar{\Sigma})$ and $\mathbf{x} = \mathbf{x}_d + \mathbf{x}_r$. This is because when both parts are generated from Gaussian distributions with the same covariance matrix, their summation is simply a Gaussian random variable with the mean vector being $\mathbf{B}_d\mathbf{u}_d + \mathbf{B}_r\mathbf{u}_r$ and covariance matrix being $\Sigma = 2\bar{\Sigma}$.

Similar to the approach I, we can derive the marginal probability of $\mathbf{x}$ from the above generative process as:

\begin{align}
&P(\mathbf{x})=\iint\limits_{\mathbf{u}_d,\mathbf{u}_r}P(\mathbf{x},\mathbf{u}_d,\mathbf{u}_r|\mathbf{B}_d,\mathbf{B}_r,\mathbf{c})\mathrm{d}\mathbf{u}_d\mathrm{d}\mathbf{u}_r \nonumber \\
&=\iint\limits_{\mathbf{u}_d,\mathbf{u}_r}P(\mathbf{x}|\mathbf{u}_d,\mathbf{u}_r,\mathbf{B}_d,\mathbf{B}_r,\mathbf{c})P(\mathbf{u}_r)P(\mathbf{u}_d|\mathbf{c})
\mathrm{d}\mathbf{u}_d\mathrm{d}\mathbf{u}_r.
\end{align}

This formulation involves an integral over latent variables $\mathbf{u}_d$ and $\mathbf{u}_r$, which makes the calculation difficult. Again,  we follow the simplification in approach I to use the point-wise maximum within the integral term to approximate the likelihood:
\begin{align}
\begin{split}
P(\mathbf{x})&\approx P(\mathbf{x}|\mathbf{u}_d^{\ast},\mathbf{u}_r^{\ast},\mathbf{B}_d,\mathbf{B}_r,\mathbf{c})P(\mathbf{u}_r^\ast)
P(\mathbf{u_d^\ast|c})\\
\mathbf{u}_d^\ast,\mathbf{u}_r^\ast &=\underset{\mathbf{u}_d,\mathbf{u}_r}{\arg\max}P(\mathbf{x}|\mathbf{u}_d,\mathbf{u}_r,\mathbf{B}_d,\mathbf{B}_r,\mathbf{c})P(\mathbf{u}_r)P(\mathbf{u_d|c}).
\end{split}
\end{align}

The negative logarithm of the likelihood is then formulated as:
\begin{align}
&-\mathrm{log}P(\mathbf{x}|\mathbf{B}_d, \mathbf{B}_r,\mathbf{c})=\min_{\mathbf{u}_d,\mathbf{u}_r}\|\mathbf{x}-\mathbf{B}_d\mathbf{u}_d-\mathbf{B}_r\mathbf{u}_r
\|_{2}^{2}+ \nonumber \\
&~~~~ \lambda_1 \|\mathbf{u}_r\|_1 + \lambda_2 \|\mathbf{u}_d\|_1 + \lambda_3\|\mathbf{u}_d-\mathbf{c}\|_{2}^2,
\label{l0_obj}
\end{align}
where the model parameters $\mathbf{B}_d$ and $\mathbf{B}_r$ can be learned by minimizing the negative logarithm of the likelihood in Eq.~(\ref{l0_obj}).

\subsection{Fisher vector derivation}
\subsubsection{Fisher vector derivation for approach I (SCFVC)}
Once the generative model is established, we can derive its Fisher vector coding for a local feature $\mathbf{x}$ by differentiating its negative log-likelihood w.r.t.\
the model parameters.

By cross-referencing the log likelihood definition of our first model in Eq. (\ref{eq:SCFVC_Log}), the Fisher vector can be calculated as follows:
\begin{align}
	\mathcal{C}(\mathbf{x}) & = \frac{\partial \log(P(\mathbf{x}|\mathbf{B}))}{\partial \mathbf{B}} = \frac{  \partial \frac{1}{\sigma^2} \|\mathbf{x}- \mathbf{B}\mathbf{u^*}\|^2_2 + \lambda \|\mathbf{u^*}\|_1}{\partial \mathbf{B}} \nonumber \\
	\mathbf{u^*} &= \mathop{\mathrm{argmax}}_{\mathbf{u}} P(\mathbf{x}|\mathbf{u},\mathbf{B}) P(\mathbf{u}).
\end{align}
Note that the differentiation involves $\mathbf{u^*}$ which implicitly interacts with $\mathbf{B}$. To calculate this term, we notice that the sparse coding problem can be reformulated as a general quadratic programming problem by defining $\mathbf{u}^+$ and $\mathbf{u}^-$ as the positive and negative parts of $\mathbf{u}$.
That is, the sparse coding problem can be rewritten as
\begin{align}
	 &\min_{\mathbf{u}^+,\mathbf{u}^-}~~ \frac{1}{\sigma^2} \| \mathbf{x} - \mathbf{B}(\mathbf{u}^+ - \mathbf{u}^-)\|^2_2 + \lambda \mathbf{1}^T(\mathbf{u}^+ + \mathbf{u}^-) \nonumber \\
   &{\rm s.t.} ~~ \mathbf{u}^+ \geq 0,~~~ \mathbf{u}^- \geq 0.
\end{align}
By further defining $\mathbf{u'} = (\mathbf{u}^+, \mathbf{u}^-)^T$, $\log(P(\mathbf{x}|\mathbf{B}))$ can be expressed in the following general form,
\begin{align}
	\log(P(\mathbf{x}|\mathbf{B})) = \mathcal{L}(\mathbf{B}) = \max_{\mathbf{u'}}~~ \mathbf{u'}^T\mathbf{v}(\mathbf{B}) - \frac{1}{2} \mathbf{u'}^T\mathbf{P}(\mathbf{B})\mathbf{u'},
\end{align}
where $\mathbf{P}(\mathbf{B})$ and $\mathbf{v}(\mathbf{B})$ are a matrix term and a vector term depending on $\mathbf{B}$ respectively. The derivative of $\mathcal{L}(\mathbf{B})$ has been studied in \cite{KernelLearning}. According to the Lemma 2 in \cite{KernelLearning}, we can differentiate $\mathcal{L}(\mathbf{B})$ with respect to $\mathbf{B}$ as if $\mathbf{u'}$ did not depend on $\mathbf{B}$. In other words, we can firstly calculate $\mathbf{u'}$ or equivalently $\mathbf{u}^*$ by solving the sparse coding problem and then obtain the Fisher vector $\frac{\partial \log(P(\mathbf{x}|\mathbf{B}))}{\partial \mathbf{B}}$ as
\begin{align}\label{eq:SC_FV}
	\frac{  \partial \frac{1}{\sigma^2} \|\mathbf{x}- \mathbf{B}\mathbf{u^*}\|^2_2 + \lambda \|\mathbf{u^*}\|_1}{\partial \mathbf{B}} = \frac{1}{\sigma^2} (\mathbf{x}- \mathbf{B}\mathbf{u^*})\mathbf{u^*}^T.
\end{align}
Note that the Fisher vector expressed in Eq. (\ref{eq:SC_FV}) has an interesting form:
{\it it is simply the outer product of
  the sparse coding vector $\mathbf{u^*}$ and the reconstruction residual term $(\mathbf{x}- \mathbf{B}\mathbf{u^*})$.
}

In traditional sparse coding, only the $k$th dimension of a coding vector $u_k$ is used to indicate the relationship between a local feature $\mathbf{x}$ and the $k$th basis. Here in Eq. (\ref{eq:SC_FV}), the coding value $u_k$ multiplying the reconstruction residual is used to capture their relationship. In the following sections, we term
this Fisher coding method Sparse Coding based Fisher vector coding (SCFVC in short).

\subsubsection{Fisher vector derivation for approach II (HSCFVC)}
Using the same technique as SCFVC, we can derive the Fisher vector coding for our second generative model:
{\footnotesize
\begin{align}
  &\mathbf{G_{B_d}^{x}} = \frac{\partial {\mathrm{log}(P(\mathbf{x}|\mathbf{B}_d,\mathbf{B}_r,\mathbf{c}))}}{\partial \mathbf{B}_d} \nonumber \\
& =\frac{\partial{\frac{1}{\sigma^2}\|\mathbf{x}-\mathbf{B}_d\mathbf{u}_d^{\ast}-\mathbf{B}_r\mathbf{u}_r^{\ast}\|_{2}^{2}+\lambda_1 \|\mathbf{u}_r^{\ast}\|_1 + \lambda_2 \|\mathbf{u}_d^{\ast}\|_1 + \lambda_3\|\mathbf{u}_d^{\ast}-\mathbf{c}\|_{2}^2}}{\partial \mathbf{B}_d} \\
&\mathbf{G_{B_r}^{x}} = \frac{\partial {\mathrm{log}(P(\mathbf{x|B_d,B_r,c}))}}{\partial \mathbf{B}_r} \nonumber \\
& =\frac{\partial{\frac{1}{\sigma^2}\|\mathbf{x}-\mathbf{B}_d\mathbf{u}_d^{\ast}-\mathbf{B}_r\mathbf{u}_r^{\ast}\|_{2}^{2}+\lambda_1 \|\mathbf{u}_r^{\ast}\|_1 + \lambda_2 \|\mathbf{u}_d^{\ast}\|_1 + \lambda_3\|\mathbf{u}_d^{\ast}-\mathbf{c}\|_{2}^2}}{\partial \mathbf{B}_r} \\
&\mathbf{u}_d^{\ast}, \mathbf{u}_r^{\ast} = \underset{\mathbf{u}_d, \mathbf{u}_r}
{\arg\min}\frac{1}{\sigma^2}\|\mathbf{x}-\mathbf{B}_d\mathbf{u}_d-\mathbf{B}_r\mathbf{u}_r\|_{2}^{2}+\lambda_1 \|\mathbf{u}_r\|_1 + \lambda_2 \|\mathbf{u}_d\|_1 \nonumber \\
& ~~~~~~~~~~~~ + \lambda_3\|\mathbf{u}_d-\mathbf{c}\|_{2}^2,
\label{FisherVector}
\end{align}}
where $\mathbf{u}_d,\mathbf{u}_r$ interact with $\mathbf{B}_d,\mathbf{B}_r$. Similar to SCFVC, we can calculate $\mathbf{G_{B_d}^{x}}$ and $\mathbf{G_{B_r}^{x}}$ as if $\mathbf{u}_d,\mathbf{u}_r$ did not depend on $\mathbf{B}_d,\mathbf{B}_r$. In other words, we can solve the inference problem in Eq. (\ref{FisherVector}) to obtain $\mathbf{u}_d^{\ast}, \mathbf{u}_r^{\ast}$ first and then calculate $\mathbf{G_{B_d}^{x}}$ and $\mathbf{G_{B_r}^{x}}$.
In the following sections, we name
this Fisher vector encoding method Hybrid Sparse Coding based Fisher vector coding (HSCFVC in short) since the creation of its final image representation involves the components of both supervised coding and Fisher vector coding.

Note that HSCFVC essentially combines two ideas of building a good classification system: 1) Identifying the discriminative pattern at the early coding stage of an image classification pipeline, i.e. supervised coding; and 2) preserving as much information of local features as possible into the high-dimensional image representation and relies on classifier learning to identify the discriminative pattern.

\begin{algorithm*}[ht!]
\caption{Matching Pursuit based algorithm for solving for  $\mathbf{u}_d,\mathbf{u}_r$ in Equation~(\ref{mp_obj})}
\begin{algorithmic}[1]
  \Procedure{$\mathbf{MP}$ }{} (see Section \ref{sec:append} for details)\\
\textbf{Input:} $\mathbf{x}, \mathbf{B}_d, \mathbf{B}_r, k_1, k_2, \lambda, \mathbf{c}$\\
\textbf{Output:} $\mathbf{u}_d,\mathbf{u}_r$ \\
Initialize residue $\mathbf{r}=\mathbf{x}, \mathbf{u}^1_d=\mathbf{0}, \mathbf{u}^1_r=\mathbf{0} $\\
Fixing $\mathbf{u}_r$, solve for $\mathbf{u}_d$
\For {$t = 1 : k_1$}
\State Solve $\min_{\mathbf{e}_{d_j},{u_d}_j}\|\mathbf{x}-\mathbf{B}_d\mathbf{u}_d^t-\mathbf{B}_r\mathbf{u}_r^t -\mathbf{B}_d \mathbf{e}_{d_j}u_{d_j}\|_{2}^{2}+ \lambda\|\mathbf{u}_d^t-\mathbf{c} + \mathbf{e}_{d_j}u_{d_j}\|_{2}^2$
\State Update $\mathbf{r} \leftarrow \mathbf{r} - \mathbf{B}_d \mathbf{e}^*_{d_j}u^*_{d_j}~$         ,  $~~~~\mathbf{u}^{t+1}_d = \mathbf{u}^t_d + \mathbf{e}^*_{d_j}u^*_{d_j}$
\EndFor \\
Fixing $\mathbf{u}_d$,  solve for     $\mathbf{u}_r$
\For {$t=1:k_2$}
\State Solve $\min_{\mathbf{e}_{r_j},{u_r}_j}\|\mathbf{x}-\mathbf{B}_d\mathbf{u}_d^t-\mathbf{B}_r\mathbf{u}_r^t
 - \mathbf{B}_r\mathbf{e}_{r_j}u_{r_j}\|_{2}^{2}$
\State Update $\mathbf{r} \leftarrow \mathbf{r} - \mathbf{B}_r \mathbf{e}^*_{r_j}u^*_{r_j}~$, $~~~~\mathbf{u}^{t+1}_r = \mathbf{u}^t_r + \mathbf{e}^*_{r_j}u^*_{r_j}$
\EndFor
\EndProcedure
\end{algorithmic}
\label{Algorithm1}
\end{algorithm*}

\subsection{Learning and inference}
To learn the model parameters and calculate the Fisher vector, we need to solve the optimization problems in Eq.~(\ref{eq:SCFVC_Log}) and Eq. (\ref{l0_obj}). These two problems can be solved using existing sparse coding solvers. However, it can  still  be
slow for problems with high-dimensional local features in practice.
In \cite{MatchingPursuit}, it has been suggested that a matching pursuit algorithm can be adopted as a substitute for  sparse coding problems
in local feature encoding approaches. Thus, in this work we use the method in \cite{MatchingPursuit} to approximately solve Eq.~(\ref{eq:SCFVC_Log}).

We also develop a similar algorithm to approximately solve Eq. (\ref{l0_obj}) which essentially solves the following variant problem of Eq.~(\ref{l0_obj}):
\begin{align}
\begin{split}
&\min_{\mathbf{u}_d,\mathbf{u}_r} \|\mathbf{x}-\mathbf{B}_d\mathbf{u}_d-\mathbf{B}_r\mathbf{u}_r\|_{2}^{2}+ \lambda\|\mathbf{u}_d-\mathbf{c}\|_{2}^2\\
&{\rm s.t.}\;\; \|\mathbf{u}_d\|_{0}\leq k_1, \quad \|\mathbf{u}_r\|_{0} \leq k_{2}.
\end{split}
\label{mp_obj}
\end{align}
In the matching pursuit algorithm, the Eq. (\ref{mp_obj}) is sequentially solved by updating one dimension of $\mathbf{u}_d$ and $\mathbf{u}_r$ at each iteration while keeping the values at other dimensions fixed. In our solution, we first update each dimension of $\mathbf{u}_d$ and then update $\mathbf{u}_r$. The algorithm is described in Algorithm \ref{Algorithm1}. For the derivation and more details of Algorithm \ref{Algorithm1}, please refer to the Appendix section.

\textcolor{black}{
To learn the model parameters $\mathbf{B}$ in SCFVC, or $\mathbf{B}_d$ and $\mathbf{B}_r$ in HSCFVC, we employ an alternating algorithm which iterates between the following two steps: 1) fixing $\mathbf{B}$ in SCFVC, or $\mathbf{B}_d$ and $\mathbf{B}_r$ in HSCFVC, then solving for
$\mathbf{u}$, or $\mathbf{u}_d$ and $\mathbf{u}_r$ in HSCFVC; 2) fixing $\mathbf{u}$, or $\mathbf{u}_d$ and $\mathbf{u}_r$ in HSCFVC, then updating $\mathbf{B}$, or $\mathbf{B}_d$ and $\mathbf{B}_r$ in HSCFVC  using
the solver proposed in \cite{FeatureSign}.}

\subsection{Implementation details}
\subsubsection{Local features}\label{sect:deepCNN_Feature}
Using the neural  activations of a pre-trained CNN model as local features has become popular recently \cite{CNN_Regional,TextureFV,Our_NIPS,SemanticFisherVector}. The local feature can be either extracted from a
fully-connected layer or a convolutional layer.
For the former case, a number of image regions are firstly sampled and each of them
passes  through the deep CNN\footnote{A faster and equivalent implementation is to convert the fully-connected layer to the convolutional layer to perform the local feature extraction process \cite{AccelerateFV}.} to extract the fully-connected layer activations which will be used as a local feature. For the latter case, the whole image is directly fed into a pre-trained CNN and the activations at each spatial location of a convolutional layer are extracted as a local feature \cite{CrossLayerPooling}. It has been observed that the fully-connected layer feature is useful for generic object classification and the convolutional layer feature is useful for texture and fine-grained image classification (the discriminative patterns are usually special types of textures). In this work, we use both types
of local features in our experiment.

\subsubsection{Pooling and normalization}
From the i.i.d.\ assumption in Eq. (\ref{eq:FisherKernel}), the Fisher vector of the whole image equals to
\begin{align}
	\frac{\partial \log(P(\mathbf{X}|\mathbf{B}))}{\partial \mathbf{B}} = \sum_{i}
  \frac{\partial \log(P(\mathbf{x}_i|\mathbf{B}))}{\partial \mathbf{B}}.
\end{align}
This is equivalent to perform the sum-pooling for the extracted Fisher coding vectors. However, it has been observed \cite{ImprovedFV,All_about_VLAD} that the image signature obtained by using sum-pooling tends to over-emphasize the information from the background \cite{ImprovedFV} or bursting visual words \cite{All_about_VLAD}. It is important to apply some normalization operations when sum-pooling is used. In this paper, we apply the intra-normalization \cite{All_about_VLAD} to normalize the pooled Fisher vectors. For example, in SCFVC we apply $l_2$ normalization to the subvectors $\sum_{i} (\mathbf{x}_i - \mathbf{B}\mathbf{u}_i^*){u_{i,k}^*} ~\forall k$, where $k$ indicates the $k$th dimension of the sparse coding $\mathbf{u}_i^*$. Besides intra-normalization, we also apply
the power normalization as suggested in \cite{ImprovedFV}.

\subsubsection{Supervised coding}\label{sect:supervised_encoding}

A wide range of supervised coding methods can be adopted in the proposed HSCFVC. However, in this paper, we only consider a particular one of them. Specifically, we encode a local feature $\mathbf{x}$ by using the following encoder:
\begin{align}
	\mathbf{c} = {f}(\mathbf{P}^T\mathbf{x}+\mathbf{b}),
\end{align}where $\mathbf{c}$ is the coding vector and $f(\cdot)$ is a nonlinear function. Here we use the soft-threshold (or hinge) function $f(a) = \max(0,a)$ as suggested in \cite{AndrewNgComparison}. The final image representation is obtained by performing sum-pooling over the coding vectors of all local features\footnote{We also apply the power-normalization in order to be consistent with the proposed SCFVC and HSCFVC.}.
To learn the encoder parameters, we feed the image representation into a logistic regression module to calculate the posterior probability and employ negative entropy as the loss function. Then $\mathbf{P}$ and $\mathbf{b}$ are jointly learned with the parameters in the logistic regressor in an end-to-end fashion through stochastic gradient descent. Note that this supervised encoder learning process is similar to performing fine-tuning on the last few layers of a convolutional neural network with $\mathbf{x}$ being the activations of a CNN\footnote{In fact, the performance of this approach is comparable with that of fine tuning a CNN network.}.

\section{Experiments}
To evaluate the effectiveness of the two proposed compositional FVC approaches, we conduct experiments on three large datasets: Caltech-UCSD Birds-200-2011 (Birds-200 in short), MIT indoor scene-67 (MIT-67) and Pascal VOC 2007 (Pascal-07).
These three datasets are commonly used evaluation benchmarks for fine-grained image classification, scene classification and object recognition.

The focus of our experiments is to verify two aspects:
1) whether the proposed SCFVC outperforms the traditional GMM based FVC (GMM-FVC);
2) whether the proposed HSCFVC outperforms SCFVC and its guiding supervised coding method in Section \ref{sect:supervised_encoding} (denoted as SupC in the following part) since HSCFVC is expected to enjoy the merits of both SupC and SCFVC\footnote{Code of SCFVC and HSCFVC is available at \url{https://bitbucket.org/chhshen/fishercoding}.}.

\begin{table*}[ht!]
\caption{Comparison of results on Birds-200. The lower part of this table lists some results in the literature.}
\centering
    \scalebox{1}
    {
\begin{tabular}{lcl}
\hline\noalign{\smallskip}
Methods & Classification Accuracy & Comments \\
\noalign{\smallskip} \hline \noalign{\smallskip}
HSCFVC (proposed) & \textbf{80.8\%} &  \\
SCFVC  (proposed) & 77.3\% &  \\
GMMFVC & 70.1\% &  \\
SupC  & 69.5\% & \\
CNN-Jitter & 63.6\% & \\
\noalign{\smallskip}\hline\noalign{\smallskip}
CrossLayer \cite{CrossLayerPooling} & 73.5\% & using convolutional features, combine two resolutions \\
                 GlobalCNN-FT \cite{ArxivNewBaseline}                  &  66.4 \% & no parts, fine tunning \\
            	 Parts-RCNN-FT    \cite{ZhangNingECCV}                  & 76.4 \% & using parts, fine tunning \\
            	 Parts-RCNN    \cite{ZhangNingECCV}                  & 68.7 \% & using parts, no fine tunning \\
			 CNNaug-SVM    \cite{CNN_Baseline}         	& 61.8\%	  &     \\
             CNN-SVM  \cite{CNN_Baseline}  			    & 53.3\%     & CNN global \\
			 DPD+CNN \cite{Decaffe}   & 65.0\%          & using  parts \\
			 DPD   \cite{Zhang_2013_ICCV}    	            & 51.0\%		     &    \\
			 Bilinear CNN   \cite{BilinearCNN}    	            & \bf 85.1\%		     & two networks, fine-tuning  \\
			 Two-Level Attention \cite{Two-Level_Attention}       & 77.9\%  &  \\
			 Unsupervised Part Model \cite{Unsupervised_Part_Model} & 81.0\% & \\
\noalign{\smallskip} \hline

\end{tabular}}
\label{tab_birds}
\end{table*}

\subsection{Experimental settings}

As mentioned above, we use the activations of a pre-trained CNN as the local features and activations from both the convolutional layer and the fully-connected layer are used. More specifically, we extract the fully-connected layer activations as the local feature for PASCAL-07 and MIT-67 because we empirically found that the fully connected layer activations work better for scene and generic object classification. For Birds-200, we use the convolutional activations as local features since it has been reported that convolutional layer activations lead to superior performance than the fully-connected layer activations when apply to the fine-grained image classification problem \cite{CrossLayerPooling}. Throughout our experiments, we use the vgg-very-deep-19-layers CNN model \cite{VGGVD} as the pre-trained CNN model. To extract the local features with the fully-connected layer activations, we first resize the input image into 512$\times$512 pixels and 614$\times$614 pixels. Then we extract regions of size 224x224 pixels at a dense spatial grid with the step size of 32 pixels. These local regions are fed into the deep CNN and the 4096-dimensional activations of the first fully-connected layer are extracted as local features. To extract the local features from the convolutional layer, we resize input images to 224$\times$224 pixels and 448$\times$448 pixels and then extract the convolutional feature activations from the ``conv5-4" layer as local features (in such setting, there are $14\times14+ 28\times28$ local features per image). To decouple the correlations between dimensions of CNN features and avoid the dimensionality explosion of the Fisher vector representation, for fully-connected layer features we apply PCA to reduce its dimensionality to 2000. For convolutional layer features, we do not perform dimensionality reduction but only use PCA for decorrelation.

Five comparing methods are implemented. Besides the proposed SCFVC, HSCFVC and the traditional GMM based FVC, the supervised coding method which serves as the guiding coding method for HSCFVC is also compared to verify if additional performance improvement can be achieved via our HSCFVC.

Moreover, we compare against a baseline method in \cite{CNN_Baseline,ArxivNewBaseline},
denoted as CNN-Jitter, which averages the fully-connected layer activations from several transformed versions of an input image, i.e.,
cropping the four corners and middle region of an input image. We also quote results of other methods reported from the literature for reference. However, since they may adopt different implementation details, their performance may not be directly comparable to ours.

Both the proposed methods and baseline methods involve several hyper-parameters.
Their settings are described as follows. In SCFVC, the codebook size of $\mathbf{B}$ is set to be 200. In HSCFVC, the dimensionality of $\mathbf{c}$ and the codebook size of $\mathbf{B}_d, \mathbf{B}_c$ are set to be 100. Therefore, the dimensionality of the image representation created by SCFVC and HSCFVC are identical. For GMM-FVC, we also set the number of Gaussian distributions to be 200 to make fair comparison. We employ the matching pursuit approximation to solve the inference problem in the SCFVC and HSCFVC. The sparsity of coding vector is controlled by the parameter $k$ in Eq. (\ref{mp_obj}). Both $k_1$ in HSCFVC and $k$ in SCFVC have significant influences on performance. We select $k_1$ from $\{10,20,30\}$ and $k$ from $\{10,20,30,40\}$ via cross-validation. $k_2$ is fixed to $10$ for simplicity. $\lambda$ in Eq. (\ref{mp_obj}) is fixed to be $0.5$ unless otherwise stated. Throughout our experiments, we use the linear SVM \cite{libsvm} as the classifier.

\subsection{Main results}
\textbf{Birds-200} Birds-200 is a commonly used benchmark for fine-grained image classification which contains 11788 images of 200 different bird species. The experimental results on this dataset are shown in Table \ref{tab_birds}. As can be seen, both the proposed SCFVC and HSCFVC outperform the traditional GMM-FVC by a large margin. The improvement can be as large as 10\%. This observation clearly demonstrates the advantage of using compositional mechanism for modeling local features.

Also, HSCFVC achieves better performance than SCFVC, which outperforms the latter by more than 3\%. Recall that the difference between HSCFVC and SCFVC lies in that the former further decomposes a local feature into a discriminative part and a residual part, thus the superior performance of HSCFVC clearly verifies the benefit of adopting such modeling.

To achieve this decomposition, HSCFVC uses a supervised coding method as guidance. Thus it is interesting to examine the performance relationship between HSCFVC and its guiding coding method. This comparison is also shown in Table \ref{tab_birds}. As we can see,
HSCFVC also outperforms its guiding supervised encoding by 11\%. As discussed previously, this further performance boost is expected because the supervised coding method may not be able to extract all discriminative patterns from local features and the missing information can be re-gained from the high-dimensional image signature generated by HSCFVC.

It can be seen that the CNN-Jitter baseline performs worst in comparison with all other methods. This suggests that to build image-level representation with a pre-trained CNN model it is better to adopt the CNN to extract local features rather than global features as in the CNN-Jitter baseline. Finally, by cross-referencing the recently published performance on this dataset, we can conclude that the proposed method is on par with the state-of-the-art. Note that some methods achieve better performance by adopting strategies which have not been considered here but can be readily incorporated into our method. For example, in \cite{BilinearCNN}, the CNN model is fine-tuned. We can use the same technique to improve our performance.

\textbf{MIT-67} MIT-67 contains 6700 images with 67 indoor scene categories. This dataset is very challenging because the differences between some categories are very subtle. The comparison of classification results is shown in Table~\ref{table:MIT67_Result}.

Again, we observe that the proposed HGMFVC and SCFVC significantly outperform traditional GMMFVC. The improvement from HSCFVC and SCFVC to GMM-FVC are around 7\% and 5\% respectively. In addition, the HSCFVC achieves superior performance than SCFVC and SupC. This again shows that HSCFVC is able to combine the benefit of both Fisher vector coding and supervised coding. By comparing our best performance against the results reported in the literature, we can see that our methods are comparable to the state-of-the-art results. The work in \cite{TextureFV} also employs the traditional GMM-FVC but achieves higher classification performance than ours. By examining their experimental setting, we found that their method actually extracts convolutional activations of a pre-trained CNN at 7 different scales which uses far more local features than ours. %

\begin{table*}[ht!]
\caption{Comparison of results on MIT-67. The lower part of this table lists some results in the literature.}
\centering
    \scalebox{1}
    {
\begin{tabular}{lcl}
\hline\noalign{\smallskip}
Methods & Classification Accuracy & Comments \\
\noalign{\smallskip} \hline \noalign{\smallskip}
HSCFVC (proposed)& \textbf{79.5\%} &  \\
SCFVC (proposed)& 77.6\% &  \\
GMMFVC & 72.6\% & \\
SupC  & 76.4\%  & \\
CNN-Jitter	& 70.2\% & \\
\noalign{\smallskip}\hline\noalign{\smallskip}
MOP-CNN\cite{CNN_Regional} & 68.9\% & with three scales\\
VLAD level2\cite{CNN_Regional} & 65.5\% & with the single best scale \\
CNN-SVM\cite{CNN_Baseline} & 58.4\% & using CNN on the whole image \\
Mid-level Mining\cite{Yao_Mining} &69.7\% & using three scales \\
SemanticFV\cite{SemanticFisherVector} &68.5\% & best performance for single scale\\
SemanticFV\cite{SemanticFisherVector} &72.9\% & using four scales\\
CrossLayer\cite{CrossLayerPooling} &71.5\%& combining two resolutions and use global CNN features \\
DeepTexture \cite{TextureFV} & \bf 81.7\% & 7 scales with convolutional layer activations\\
Bilinear CNN \cite{BilinearCNN} & 77.6\% & without fine-tuning \\
Bilinear CNN \cite{BilinearCNN} & 71.1\% & fine-tuning \\
\noalign{\smallskip} \hline
\end{tabular}}
\label{table:MIT67_Result}
\end{table*}

\textbf{Pascal-07} The Pascal VOC 2007 dataset is composed of 9963 images of 20 object categories. The task is to predict whether a target object is present in an image or not. Table \ref{tab_voc} shows the results measured by mean average precision (mAP) of all 20 classes. Table \ref{table:Pascal07_Each_Class} provides performance comparison on each one of the 20 categories.

As we can see, the HSCFVC and SCFVC again outperform the traditional GMM-FVC. Also, HSCFVC achieves the best classification performance. By cross-referencing Table \ref{table:Pascal07_Each_Class}, it is observed that the relative performance of HSCFVC, SCFVC and GMMFV is almost kept for all 20 classes, that is, HSCFVC always achieves the best performance and SCFVC is superior over GMMFVC.

By taking a close examination on Table \ref{table:Pascal07_Each_Class}, we observe that the proposed method usually achieves the largest improvement on the difficult categories (those categories with less than 90\% mAP), e.g.,
the categories ``TV'', ``Sheep'', ``Bottle'', ``Chair''. This can be understood by the fact that for difficult classes, many subtle class differences can only be captured by very discriminative image representations.

\textcolor{black}{
Note that the method in \cite{VGGVD} directly applies average pooling on the fully-connected layer activations extracted from local image regions and it achieves 89.3\% mAP. However, they choose a different scheme to crop image regions. Different from our setting, they maintain the aspect ratio of input images and sample across multiple scales (5 scales in total). We also experiment with a similar setting, that is, to retain the aspect ratio of input images and sample in 3 scales, and we can achieve comparable or even better performance (HGMFV achieves 89.8\% mAP). Using this setting, we also compare GMMFVC, SCFVC, HSCFVC and average pooling (the method in \cite{VGGVD} on PASCAL 2012). We train the model on the training set and evaluate the result on the validation set. The results are shown in Table \ref{table:Pascal12_Each_Class}. As can be seen, the performance relationship of different methods is consistent with that in PASCAL 2007.}

\begin{table*}[!ht]
        \caption{Comparison of results on Pascal VOC 2007 for each of 20 classes. }
        	\centering
		\label{table:Pascal07_Each_Class}
		\begin{tabular}{llllllllllll}
		\hline\noalign{\smallskip}
  & TV & train & sofa & sheep & plant & person & mbike & horse & dog & table \\
         \noalign{\smallskip}
         \hline
         \noalign{\smallskip}
Global Jitter  &  81.9 & 96.3 & 72.9 & 86.1 & 61.6 & 95.2 & 89.3 & 91.5 & 90.9 & 79.7 \\
GMMFVC &  81.3 & 95.8 & 77.3 & 80.6 & 63.2 & 95.9 & 89.9 & 92.1 & 89.1 & 79.1 \\
SCFVC  & 84.1 & 96.4 & 79.7 & 84.2 & 64.2 & 96.2 & 90.4 & 93.8 & 90.9 & 83.1 \\
HSCFVC  & 87.4 & 96.7 &  \bf 80.0 & 88.6 & 65.9 & 97.1 & 92.5 & 94.6 & 93.9 & 84.2 \\
HSCFVC (region-crop as \cite{VGGVD}) & \bf 89.4 & \bf 98.3 & 79.1 & \bf 91.8 & \bf 67.9 & \bf 97.8 & \bf 94.4 & \bf 96.3 & \bf 96.2 & \bf 84.5  \\
         \noalign{\smallskip}
         \hline
         \noalign{\smallskip}
  &cow & chair & cat & car & bus & bottle & boat & bird & bike & areo & mAP\\
           \noalign{\smallskip}
         \hline
         \noalign{\smallskip}
Global Jitter  & 77.8 & 67.4 & 92.1 & 91.2 & 85.2 & 56.6 & 92.8 & 92.9 & 90.4 & 97.1 & 84.4\\
GMMFVC  & 81.1 & 66.6 & 93.3 & 92.3 & 86.4 & 58.9 & 89.2 & 93.5 & 92.2 & 94.9 & 84.6\\
SCFVC  & 81.7 & 68.2 & 92.6 & 91.9 & 88.4 & 61.8 & 90.6 & 93.6 & 92.6 & 97.3 & 86.6\\
HSCFVC  &   83.4 & 72.2 & 95.2 & 93.9 & 90.3 & 65.0 & 92.5 & 95.3 & 94.0 & 97.6 & 88.1\\
HSCFVC (region-crop as \cite{VGGVD}) & \bf 86.8 & \bf 74.3 & \bf 96.7 & \bf 94.4 & \bf 91.5 & \bf 68.6 & \bf 95.5 & \bf 96.8 & \bf 95.7 & \bf 98.7 & \bf 89.8\\
\hline
      \end{tabular}
\end{table*}

\begin{table*}[!ht]
        \caption{Comparison of results on Pascal VOC 2012 for each of 20 classes. (region-crop as \cite{VGGVD}) }
        	\centering
		\label{table:Pascal12_Each_Class}
		\begin{tabular}{lllllllllllll}
		\hline\noalign{\smallskip}
  & TV & train & sofa & sheep & plant & person & mbike & horse & dog & table & -\\
         \noalign{\smallskip}
         \hline
         \noalign{\smallskip}
Average Pooling  &  83.4 & 96.8 & 64.8 & 89.6 & 54.0 & 95.7 & 91.2 & 91.4 & 96.1 & 73.8 \\
GMMFVC &  87.2 & 97.5 & 71.5 & 89.1 & 56.0 & 95.8 & 91.3 & 90.8 & 94.7 & 76.2 \\
SCFVC  & 87.5 & 97.2 & \bf 76.0 & 91.8 & 62.4 & \bf 96.7 & 93.0 & 92.7 & 96.3 & 79.5 \\
HSCFVC  & \bf 87.9 & \bf 97.7 &  75.8 &  \bf 92.2 &  \bf 62.6 & \bf 96.7 &  \bf 93.4 & \bf 93.6 &  \bf 96.6 & \bf 81.1 \\
         \noalign{\smallskip}
         \hline
         \noalign{\smallskip}
  &cow & chair & cat & car & bus & bottle & boat & bird & bike & areo & mAP\\
           \noalign{\smallskip}
         \hline
         \noalign{\smallskip}
Average Pooling   & 83.2 & 70.5 & 96.9 & 78.5 & 92.9 & 64.8 & 89.1 & 94.4 & 86.2 & 98.5 & 84.6 \\
GMMFVC  & 84.3 & 75.4 & 95.9 & 82.2 & 93.2 & 64.5 & 88.6 & 94.4 & 88.5 & 97.6 & 85.8\\
SCFVC  & 86.3 & 77.7 & 97.3 & 83.7 & 94.4 & 70.2 & 90.3 & 95.9 & 90.4 & 98.5 & 87.9 \\
HSCFVC  &  \bf 88.6 & \bf 78.8 &  \bf 97.4 & \bf 84.3 &  \bf 94.8 & \bf 71.2 & \bf 90.5 &  \bf 96.0 & \bf 91.0 & \bf 99.8 & \bf 88.5\\
\hline
      \end{tabular}
\end{table*}

\begin{table*}[ht!]
\caption{Comparison of results on Pascal VOC 2007. The lower part of this table lists some results in the literature.}
\centering
    \scalebox{1}
    {
\begin{tabular}{lll}
\hline\noalign{\smallskip}
Methods & Mean Average Precision & Comments \\
\noalign{\smallskip} \hline \noalign{\smallskip}
HSCFVC (proposed)& \bf 88.1\% &  \\
HSCFVC (region-crop as \cite{VGGVD}) & \bf 89.8\% \\
SCFVC (proposed) & 86.6\% &  \\
GMMFVC & 84.6\% & \\
SupC & 84.2\%  & \\
\noalign{\smallskip}\hline\noalign{\smallskip}
CNNaug-SVM\cite{CNN_Baseline} & 77.2\% & with augmented data, use CNN on whole image \\
CNN-SVM\cite{CNN_Baseline} & 73.9\% & no augmented data, use CNN on whole image \\
Deep Fisher\cite{End_to_end_deep_fisher} & 56.3\% & training GMM parameters in an end-to-end fashion \\
Mid-level Mining\cite{Yao_Mining} & 75.2\% & using three scales\\
CrossLayer\cite{CrossLayerPooling} & 77.8\% & combining two resolutions and global CNN features \\
DeepTexture \cite{TextureFV} &  84.9\% & \\
SPP-Net \cite{SPP_Net} & 82.9\% &   \\
CNN S TUNE-RNK  \cite{Devil_in_the_details} &  82.4\%  &    \\
CNN aggregation \cite{VGGVD} & \bf 89.3\% & using a different cropping scheme \\
\noalign{\smallskip} \hline
\end{tabular}}
\label{tab_voc}
\end{table*}

\begin{table*}[ht!]
        \caption{Comparison of results on MIT-67 with three different settings: (1) 200-basis codebook with 1000 dimensional local features (2) 500 Gaussian distributions with 400 dimensional local features (3) 1000 Gaussian distributions with 200 dimensional local features. They have the same total image representation dimensionality. }
		\centering
		\label{table:two_schemes_comparison_mit67}
		\begin{tabular}{lccc}
		\hline\noalign{\smallskip}
				Methods & Codebook size & Local feature dimension & Accuracy\\
				        & /Number of Gaussian distributions &           &  \\
		\noalign{\smallskip} \hline \noalign{\smallskip}
                SCFVC   & 200 & 1000 &  \bf 77.2\%\\
                GMMFVC & 200 & 1000 & 72.6\% \\
                GMMFVC  & 500 & 400 & 73.7\%\\
                GMMFVC  & 1000 & 200 &  71.1\%\\
            \hline
      \end{tabular}
\end{table*}

\begin{table*}[ht!]
        \caption{Comparison of results on Birds-200 with three different settings: (1) 200-basis codebook with 512 dimensional local features (2) 256 Gaussian distributions with 400 dimensional local features (3) 400 Gaussian distributions with 400 dimensional local features. They have similar total image representation dimensionality. }
		\centering
		\label{table:two_schemes_comparison_birds}
		\begin{tabular}{lccc}
		\hline\noalign{\smallskip}
				Methods & Codebook size & Local feature dimension & Accuracy\\
				        & /Number of Gaussian distributions &           &  \\
		\noalign{\smallskip} \hline \noalign{\smallskip}
                SCFVC   & 200 & 512 &  \bf 77.3\%\\
                GMMFVC & 200 & 512 & 70.1\% \\
                GMMFVC  & 256 & 400 & 71.1\%\\
                GMMFVC  & 400 & 256 &  74.8\%\\
                GMMFVC  & 800 & 128 & 76.3\% \\
                GMMFVC  & 2048 & 50 & 76.4\% \\
            \hline
      \end{tabular}
\end{table*}

\begin{table*}[ht!]
        \caption{The impact of the number of bases on our methods. }
		\centering
		\label{table:impact of bases }
		\begin{tabular}{lccc}
		\hline\noalign{\smallskip}
				Methods &   Number of bases &  Accuracy\\
		\noalign{\smallskip} \hline \noalign{\smallskip}
                SCFVC   & 200  & 77.6\%\\
                SCFVC   & 400  & 77.3\% \\
                SCFVC   & 600  & 77.2\% \\
                HSCFVC  & 200  & 79.5\%\\
                HSCFVC  & 400  & 79.4\%\\
                HSCFVC  & 600  & 78.9\%\\
            \hline
      \end{tabular}
\end{table*}

\subsection{Analysis of SCFVC}
\subsubsection{GMMFVC vs. SCFVC: the impact of local feature dimensions}

\begin{figure}[ht]
\begin{center}
\includegraphics[scale=0.5]{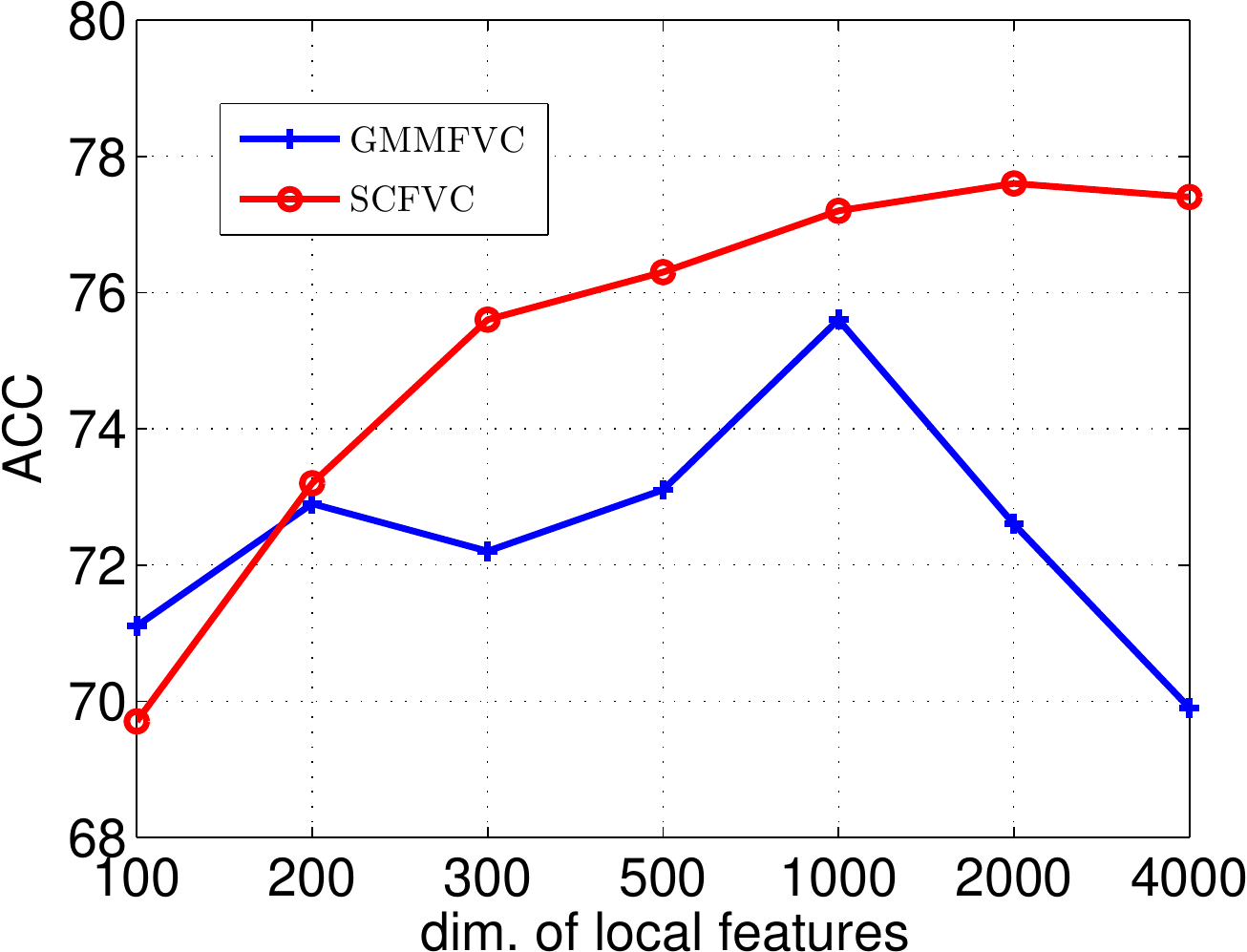}
\end{center}
\caption{Comparison of GMMFVC and SCFVC with different local feature dimensionality. The experiment is conducted on the MIT-67 dataset.}
\label{fig_gmm_scfv}
\end{figure}

In the above experiments, the dimensionality of local features is fixed to 2000. Compared with the traditional setting of GMMFVC, e.g. with 128 dimensional SIFT feature, its dimensionality is relatively high.
How about the performance comparison between the proposed SCFVC and traditional GMMFVC on lower dimensional features? To investigate this issue, we vary the dimensionality of the deep CNN feature from 100 to 2000 and compare the performance of the two Fisher vector coding methods on MIT-67. The results are shown in Figure \ref{fig_gmm_scfv}. From Figure \ref{fig_gmm_scfv}, we could make several interesting observations. \begin{itemize}
	\item  \textcolor{black}{When the dimensionality of local feature is low, e.g., the dimensionality being 100, the performance of SCFVC and traditional GMMFVC are comparable. GMMFVC even tends to perform better than SCFVC if the local feature dimensionality is low. In fact, when try lower local feature dimensions, e.g., 50, SCFVC only achieves 62\% classification accuracy while GMMFVC achieves 69\%.}
	\item In general, with the increase of local feature dimensionality, performance improved can be observed with both methods. However for the traditional GMMFVC, the performance gain obtained from increasing feature dimensionality is lower than that obtained by the proposed SCFVC.
	\item For traditional GMMFVC, worse performance can even obtained with further increase of local feature dimensionality.
\end{itemize}

From the above observation, we can conclude that SCFVC is more suited for encoding high dimensional local features.

\subsubsection{GMMFVC vs.\ SCFVC: codebook size and feature dimensionality trade-off}

Since GMMFVC works well for lower dimensional feature, then how about reducing the higher dimensional local feature to lower dimensions and use more Gaussian distributions? Will it be able to achieve comparable performance to our SCFVC which uses higher dimensional local features but smaller number of bases? To investigate this issue, we compare different combinations of codebook size and feature dimensionality
for GMMFVC.

We conduct our experiment on the MIT-67 and Birds-200 datasets.
The former uses the fully-connected layer activations as local features while the latter uses the convolutional layer activations as local features. For both datasets, we vary the feature dimensionality through PCA and for each test dimensionality we choose a codebook size which makes the total dimensionality of Fisher vectors be identical to that in the SCFVC baseline. The comparison results are shown in Table \ref{table:two_schemes_comparison_mit67} and Table \ref{table:two_schemes_comparison_birds}.

As we observe from both tables,  all the variants of GMM-FVC are still inferior to SCFVC.
For GMM-FVC, we do observe performance improvement by moderately reducing feature dimensionality but increasing codebook size. For example, for MIT-67, when the codebook size increases to 500, the performance of GMM-FVC improves 1\%; for Birds-200, when the codebook size increases to 800, the performance of GMM-FVC improves 6.3\%. However, further increasing codebook size and reducing local feature dimensionality does not lead to further improvement. From the results in MIT-67, obvious performance drop is even observed. This suggests that some discriminative information may have already been lost after the PCA dimensionality reduction and the discriminative power can not be re-boosted by simply introducing more Gaussian distributions. This verifies the necessity of using high dimensional local feature and justifies the value of the proposed method.

\textcolor{black}{We also examine the impact of the number of bases in our methods. We conduct the experiment on MIT67 and the results are shown in Table \ref{table:impact of bases }. Clearly, for both SCFVC and HSCFVC, their performance is not very sensitive to the number of bases. }

\subsection{Analysis of HSCFVC}

  \begin{figure*}[!ht]
  	\centering
    \begin{tabular}{l}
            \subfloat[]{ \includegraphics[height=50mm,width=57mm]{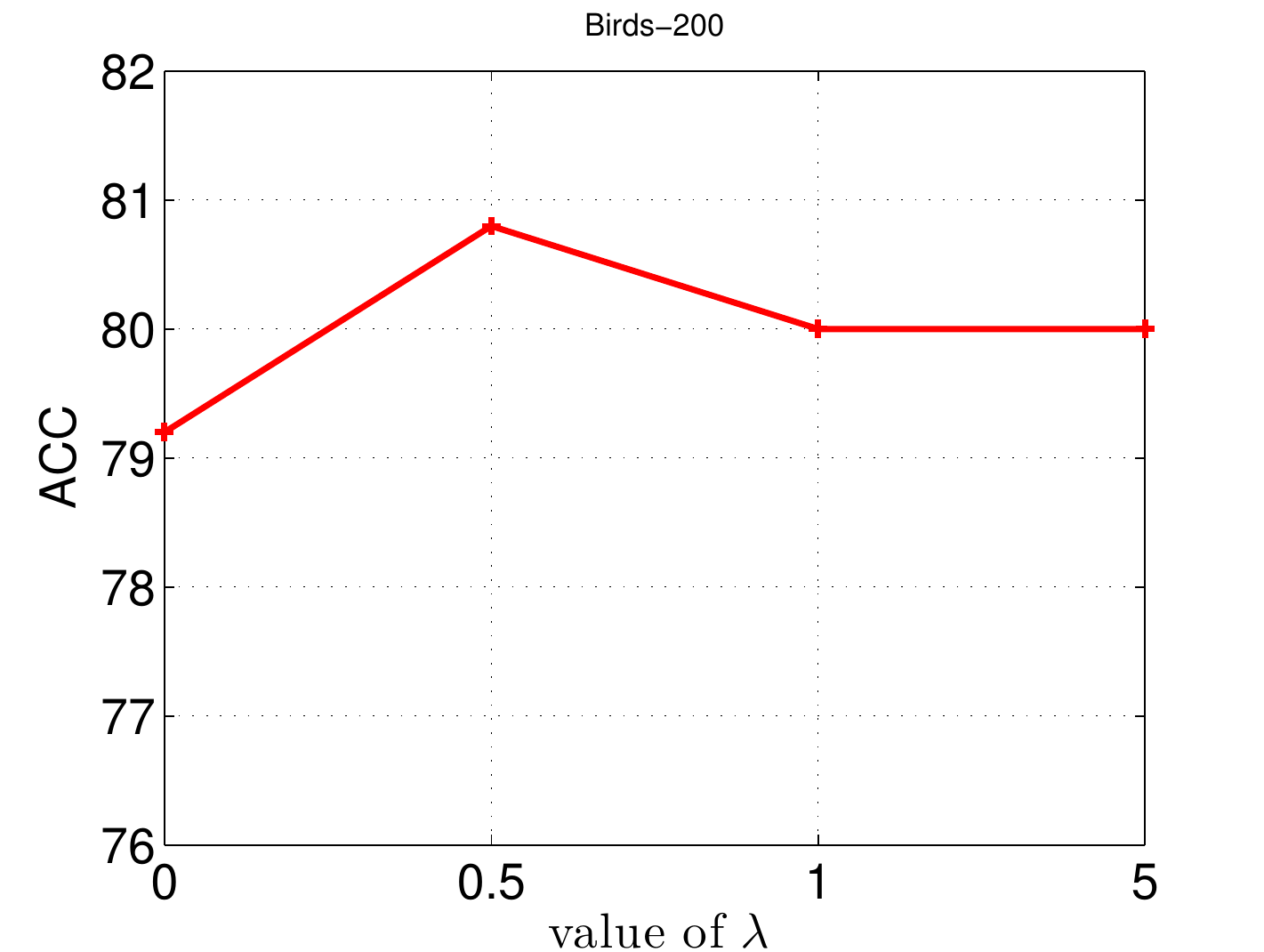}}
            \subfloat[]{ \includegraphics[height=50mm,width=57mm]{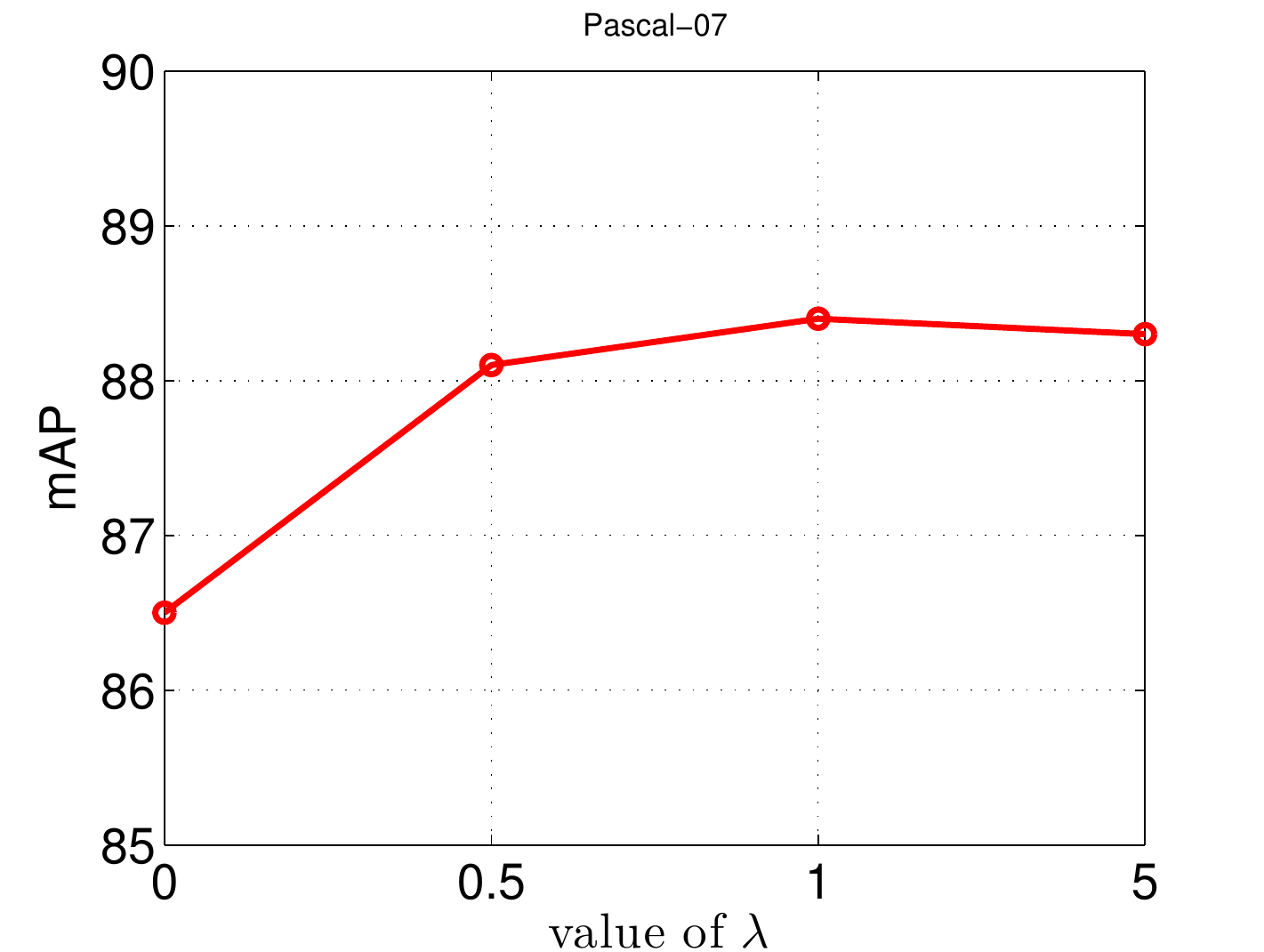}}
            \subfloat[]{ \includegraphics[height=50mm,width=57mm]{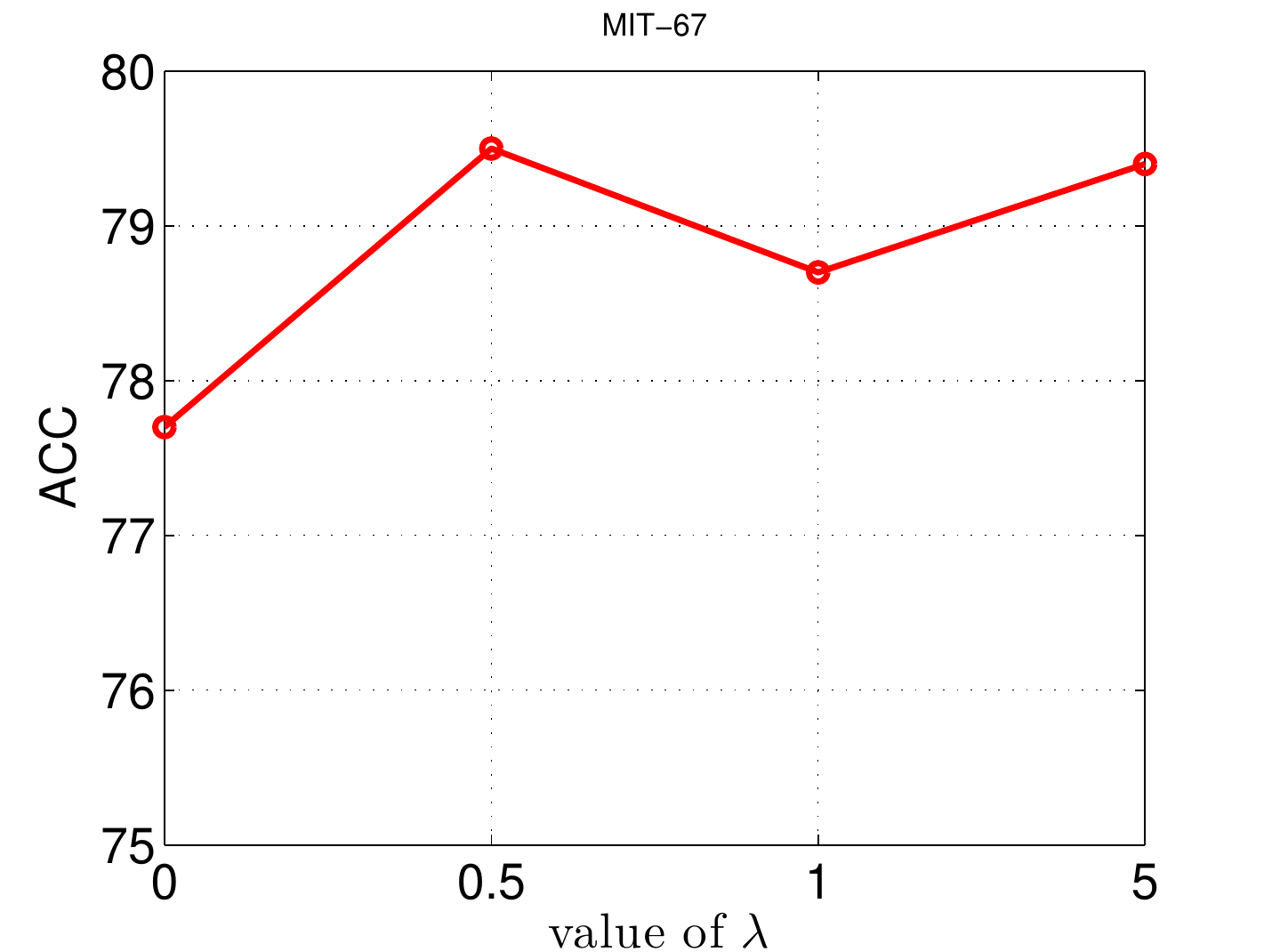}}
    \end{tabular}
    \caption{ The impact of the parameter $\lambda$ in Eq. (\ref{mp_obj}) on the classification performance. (a) result on Birds-200 (b) result on Pascal-07 (c) result on MIT-67.
    }
    \label{fig:fig_lambda}
    \end{figure*}

\begin{figure}[ht]
\begin{center}
\includegraphics[height=50mm,width=80mm]{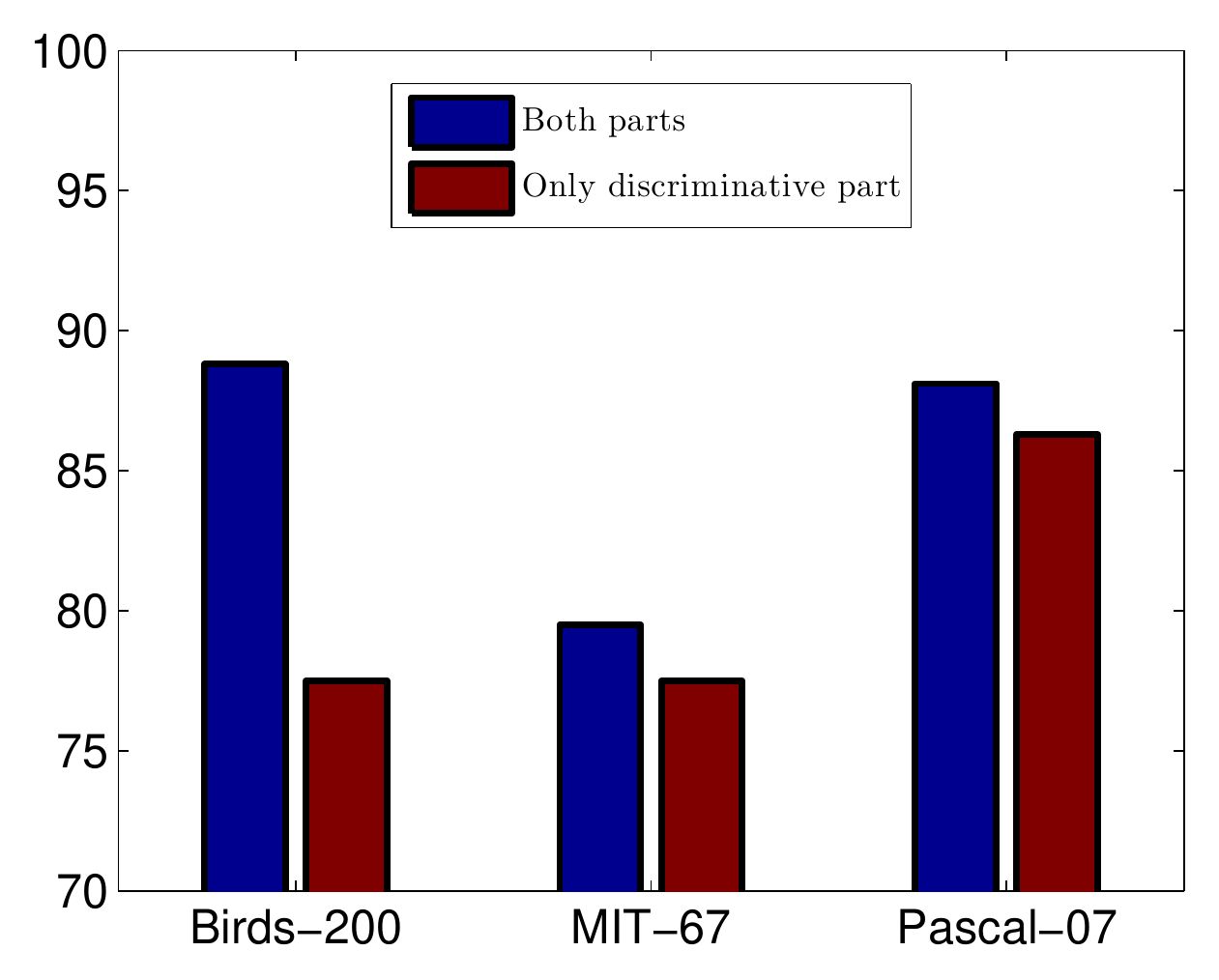}
\end{center}
\caption{Comparison of with and without the common part Fisher vector.}
\label{fig_two_parts}
\end{figure}
\subsubsection{The classification accuracy vs. the value of $\lambda$}

In HSCFVC, the optimal coding vector is calculated by solving Eq. (\ref{mp_obj}). The optimization in Eq. (\ref{mp_obj}) involves a trade-off parameter $\lambda$ which controls the fidelity of $\mathbf{u}_d$ to the supervised coding vector $\mathbf{c}$. Larger $\lambda$ enforces the active elements of $\mathbf{u}_d$ to be consistent with those of $\mathbf{c}$.
While on the contrary, smaller $\lambda$ losses this consistency.

In this subsection, we evaluate its impact on the classification performance. We conduct our experiment on all three datasets and the results are shown in Figure \ref{fig:fig_lambda}. As can be seen, for all datasets, it leads to poor performance  if $\lambda$ is set to 0.
This is not surprising because in this case the guidance signal of the supervised coding method is completely disabled.

As expected, when $\lambda$ increases, the classification accuracy rises accordingly. The performance becomes steady when $\lambda$ is reasonably large. This suggests the necessity of introducing the fidelity term $\|\mathbf{u_d - c}\|^2_2$ in Eq. (\ref{mp_obj}). Also, as can be seen, as long as $\lambda$ is sufficient large, the classification performance does not vary too much with the choice of $\lambda$ and that is why we simply set $\lambda$ to 0.5 throughout our experiments.

\subsubsection{The impact of the residual part Fisher vector $\mathbf{G_{B_c}^{X}}$ on classification performance}

Recall that the Fisher vector in HSCFVC can be decomposed into two parts: the discriminative part $\mathbf{G_{B_d}^{X}}$ and the residual part $\mathbf{G_{B_c}^{X}}$. By default we use both parts for classification because we postulate that these two parts can compensate each other. In this subsection, we verify this point by comparing the performance of merely using $\mathbf{G_{B_d}^{X}}$, the discriminative part Fisher vector and using both parts.

Note that we do not compare the scheme of only using the residual part here because only using the residual part is equivalent to SCFVC which has already been compared in the previous sections. The comparison results are shown in Figure \ref{fig_two_parts}.
We see that, for all three datasets, combining $\mathbf{G_{B_c}^{X}}$ and $\mathbf{G_{B_d}^{X}}$ can lead to better performance, which supports our assumption.

\section{Conclusion}
In this paper, we propose a new methodology of building generative models for deriving Fisher vector coding. Our key idea is to adopt a compositional mechanism into the generative process of local features. With this mechanism, the local feature could be sampled from a Gaussian distribution whose mean vector can be composed of a set of bases rather than being chosen from a fixed number of mean vectors as in the traditional GMM based Fisher vector. Based on this idea, we develop two Fisher vector coding methods. The first one adopts a single basis matrix to model local features while the second one adopts two bases to separately model the discriminative and residual parts of local features. For the second method, a guiding supervised coding method is also utilized and the second method inherits the merits of both supervised coding and Fisher vector coding. Throughout our experimental evaluation, we conclude that both methods outperform the traditional GMM based Fisher vector coding while our second method, hybrid sparse coding based Fisher vector coding, achieves the overall best performance.

\section{Matching pursuit based optimization for Equation (\ref{mp_obj})}
\label{sec:append}

Here we describe, in detail, a matching pursuit based method to solve the optimization problem Eq.~(\ref{mp_obj}).
Given the bases $\mathbf{B}_d \in \mathbb{R}^{d\times m_1}$ and $\mathbf{B}_r \in \mathbb{R}^{d\times m_2}$,
matching pursuit sequentially updates one dimension of $\mathbf{u}_d$ (or $\mathbf{u}_r$) at each iteration in order to minimize the objective function
in Eq.~(\ref{mp_obj}).
In practice, we firstly solve for $\mathbf{u}_d$ and  $\mathbf{u}_r$ alternatively. The optimization problem at each iteration is as follows.

Solving for $\mathbf{u}_d$:
\begin{align}
\begin{split}
&\min_{\mathbf{e}_{d_j},{u}_{d_j}}\|\mathbf{x}-\mathbf{B}_d\mathbf{u}_d^t-\mathbf{B}_r\mathbf{u}_c^t
 - \mathbf{B}_d \mathbf{e}_{d_j}u_{d_j}\|_{2}^{2}+ \\
 & ~~~~~~~~~~~~~ \lambda\|\mathbf{u}_d^t-\mathbf{c} + \mathbf{e}_{d_j}u_{d_j}\|_{2}^2\\
\end{split}
\label{2}
\end{align}

Solving for $\mathbf{u}_r$:
\begin{align}
\begin{split}
&\min_{\mathbf{e}_{r_j},u_{r_j}}\|\mathbf{x}-\mathbf{B}_d\mathbf{u}_d^t-\mathbf{B}_r\mathbf{u}_r^t
 - \mathbf{B}_r\mathbf{e}_{r_j}u_{r_j}\|_{2}^{2},\\
\end{split}
\label{3}
\end{align}
where $\mathbf{u}_d^t$ and $\mathbf{u}_c^t$ are the solutions for $\mathbf{u}_d$ and $\mathbf{u}_r$ at the $t$-th iteration respectively. $\mathbf{e}_{d_j}$ and $\mathbf{e}_{r_j}$ are binary vectors with only one nonzero entry at the $j$th dimensions, $j \in [0,1,\cdots,m_1]$ for $\mathbf{e}_{d_j}$ and $j \in [0,1,\cdots,m_2]$ for $\mathbf{e}_{r_j}$. Therefore there are $m_1$ possible choices for $\mathbf{e}_{d_j}$ and $m_2$ possible choices for $\mathbf{e}_{r_j}$ respectively. They indicate which dimension of $\mathbf{u}_d$ ($\mathbf{u}_r$) is to be updated and the scalar $u_{d_j}$ ($u_{r_j}$) denotes the value to be updated at the chosen dimension.

To solve for $\mathbf{e}_{d_j}$ ($\mathbf{e}_{r_j}$), we simply test its all possible choices and for each candidate $\mathbf{e}_{d_j}$ ($\mathbf{e}_{r_j}$) its corresponding optimal $\bar{u}_{d_j}$ ($\bar{u}_{r_j}$) can be analytically calculated:
\begin{align}
\begin{split}
&\bar{u}_{d_j}=\frac{\mathbf{r}^{\top}\mathbf{B}_{d_j}+\lambda{c_j}}
{\mathbf{B}_{d_j}^{\top}\mathbf{B}_{d_j}+\lambda},\\
&\bar{u}_{c_j}=\frac{\mathbf{r}^{\top}\mathbf{B}_{c_j}}
{\mathbf{B}_{r_j}^{\top}\mathbf{B}_{c_j}},
\end{split}
\label{6}
\end{align}
where $\mathbf{B}_{d_j} = \mathbf{B}_d \mathbf{e}_{d_j}$ ($\mathbf{B}_{r_j} = \mathbf{B}_r \mathbf{e}_{r_j}$) is the $j$th column of $\mathbf{B}_{d_j}$ ($\mathbf{B}_{r_j}$).

Thus the objective value in Eq.~(\ref{2}) (and Eq.~(\ref{3})) can be calculated by substituting $\mathbf{e}_{d_j}$ ($\mathbf{e}_{c_j}$)
and its corresponding $\bar{u}_{d_j}$ ($\bar{u}_{r_j}$).
We select the best $\mathbf{e}_{d_j}$ ($\mathbf{e}_{r_j}$) and its corresponding $\bar{u}_{d_j}$ ($\bar{u}_{r_j}$) which minimize
Eq.~(\ref{2}) (Eq. (\ref{3})) as the solution, denoted by
 $\mathbf{e}^*_{d_j}$ ($\mathbf{e}^*_{r_j}$) and ${u}^*_{d_j}$ (${u}^*_{r_j}$).

Then $\mathbf{u}_d,\mathbf{u}_r$ can be updated as $\mathbf{u}^{t+1}_d = \mathbf{u}^t_d + \mathbf{e}^*_{d_j}u^*_{d_j}$ and $\mathbf{u}^{t+1}_r = \mathbf{u}^t_r + \mathbf{e}^*_{r_j}u^*_{r_j}$.
To avoid redundant computation, we
define a residual term $\mathbf{r} =\mathbf{x}-\mathbf{B}_d\mathbf{u}_d^t-\mathbf{B}_r\mathbf{u}_r^t$  and update it by $\mathbf{r} \leftarrow \mathbf{r} - \mathbf{B}_d \mathbf{e}^*_{d_j}u^*_{d_j}$ and $\mathbf{r} \leftarrow \mathbf{r} - \mathbf{B}_r \mathbf{e}^*_{c_j}u^*_{c_j}$.

\section*{Acknowledgements}

C. Shen's participation was in part supported by Australian Research Council Future Fellowship (FT120100969).
H. T. Shen's participation was in part supported by
National Nature Science Foundation of China
(61632007).

L. Liu and P. Wang contributed equally to this work. C. Shen is the corresponding author.

\bibliographystyle{IEEEtran}
\bibliography{CSRef}

\end{document}